# Topic Classification of Case Law Using a Large Language Model and a New Taxonomy for UK Law: AI Insights into Summary Judgment

Holli Sargeant,[*†] Ahmed Izzidien[*] and Felix Steffek[‡]

## Abstract

This paper addresses a critical gap in legal analytics by developing and applying a novel taxonomy for topic classification of summary judgment cases in the United Kingdom. Using a curated dataset of summary judgment cases, we use the Large Language Model Claude 3 Opus to explore functional topics and trends. We find that Claude 3 Opus correctly classified the topic with an accuracy of 87.13% and an F1 score of 0.87. The analysis reveals distinct patterns in the application of summary judgments across various legal domains. As case law in the United Kingdom is not originally labelled with keywords or a topic filtering option, the findings not only refine our understanding of the thematic underpinnings of summary judgments but also illustrate the potential of combining traditional and AI-driven approaches in legal classification. Therefore, this paper provides a new and general taxonomy for UK law. The implications of this work serve as a foundation for further research and policy discussions in the field of judicial administration and computational legal research methodologies.

* Faculty of Law, University of Cambridge, Cambridge, United Kingdom.

† Corresponding author: hs775@cam.ac.uk.

‡ Faculty of Law, University of Cambridge, Cambridge, United Kingdom; University of Notre Dame Law School.

This work was funded by the Nuffield Foundation grant Access to Justice Through Artificial Intelligence. The views expressed are those of the authors and not necessarily of the Foundation. Visit www.nuffieldfoundation.org.

This version of the article has been accepted for publication, after peer review but is not the Version of Record and does not reflect post-acceptance improvements, or any corrections. The Version of Record is available online and please cite as:

Holli Sargeant, Ahmed Izzidien and Felix Steffek, 'Topic Classification of Case Law Using a Large Language Model and a New Taxonomy for UK Law: AI Insights into Summary Judgment' (2025) *Artificial Intelligence and Law*. http://dx.doi.org/10.1007/s10506-025-09434-0.

# 1 Introduction

Case law in the United Kingdom (**UK**) is not labelled with keywords or topic classifications, meaning it can be challenging to research or analyse issues across a large volume of legal information. In various fields where large textual datasets prevail, topic modelling has proven to be an invaluable tool, offering insights into hidden thematic structures that underpin vast corpora. However, its application in the legal domain, particularly in analysing case law, remains underexplored. This paper introduces a pioneering approach to computational legal analysis by developing a functional taxonomy of legal areas and conducting legal area classification of summary judgment cases in the UK.

Summary judgment is an important procedure used across almost all civil law procedures and courts, making it an ideal ground for topic classification. It allows courts to resolve cases efficiently without a full trial when there is "no real prospect" of success at trial and no other "compelling reason" for a trial.[1] Although they are a key procedural tool in civil litigation, it is often used against self-represented litigants who may not have the legal knowledge to effectively defend themselves (Leitch 2018; Halligan 2022). The increasing use of summary judgments raises important concerns about balancing efficiency and fairness while also maintaining access to justice (Bronsteen 2007; Gelbach 2014). The aim of this study is to identify patterns and trends in the use of summary judgment to provide a clearer picture of its role in UK cases.

Despite the extensive application of topic modelling in other domains, the application of traditional topic model methods to legal data faces significant challenges. These methods often fail to capture the complex and nuanced nature of legal language and reasoning, underscoring a critical gap in current research methodologies. We test whether recent developments in large language models (**LLMs**) offer researchers a new avenue to undertake topic modelling, or more accurately, topic classification. Using this novel method, we prompt an LLM to classify the topic of a court case by providing it with a taxonomy from which it must be selected, providing the full text of the case, and implementing a simple self-evaluation on the LLM response to improve its replies.

Our study addresses the following research questions:

1. How should a functional taxonomy of legal topic areas be defined?
2. Can LLMs effectively and accurately classify summary judgment case law by topic?
3. Can LLM-based topic classification reveal themes and patterns within case law?

To answer these questions, we develop a new functional taxonomy of legal area topics and employ advanced techniques using LLM Claude 3 Opus (Anthropic 2024) over a dataset of 3078 UK summary judgment cases. The significance of this study lies in its dual contribution to both legal practice and academic research. By providing a detailed analysis of how summary judgment cases are distributed across time and legal subject matter, we offer unique insights that can influence future legal analyses and policymaking. Furthermore, this work lays a foundational methodology for subsequent research in computational legal studies, particularly in the development and application of AI-driven tools for legal document analysis.

Our contributions are as follows:

1. We have developed a functional taxonomy for legal topics tailored to the UK legal system and used it in our application of LLMs to classify legal corpora.

---

[1] Civil Procedure Rules, pt 24.

2. We present the first study on legal area classification of summary judgments and the first paper to use an LLM to topic model legal corpora with a bespoke functional legal taxonomy. We demonstrate a high accuracy achieved with this method (87.13% overall).
3. Our analysis reveals critical trends and shifts in the application of the summary judgment procedure across time, courts and topics, offering valuable perspectives for legal scholars, practitioners and policymakers.
4. The methodology and findings presented here, including our taxonomy and prompt, lay a foundational framework for future research in legal AI. This includes the potential expansion of legal area classification techniques to other areas of law and the refinement of LLM applications in legal contexts.

The remainder of this paper is structured as follows. Section 2 reviews the most recent literature on topic modelling in the legal domain. Section 3 introduces a new functional legal taxonomy for UK law. Section 4 details the curated dataset and describes the methodologies used for legal area classification, including approaches using the LLM, Claude Opus 3. Section 5 presents the outcomes of the legal area classification, offering insights into thematic patterns within summary judgment cases across time and courts. Section 6 interprets the findings, discusses their implications for both legal practice and academic research, and considers how these insights might influence future legal analyses and policymaking. Section 7 concludes.

## 2 Related Work

Topic modelling is a computational method for identifying topics within a large dataset. Traditional topic model methods involve a model grouping similar words into topics based on their distribution across a large volume of text. This usually requires a human expert to manually allocate topics to the 'bag of words' identified by the model (Chauhan and Shah 2021). Despite being a well-established method in computer science, it has only been recently adopted in legal research. Several papers have used traditional topic model methods, including Latent Dirichlet Allocation (**LDA**), Non-Negative Matrix Factorisation (**NMF**) and Latent Semantic Analysis (**LSA**), to analyse legal texts (Lu et al. 2011; Venkatesh and Raghuveer 2012; O'Neill 2016; Remmits 2017; Luz De Araujo and De Campos 2020; Izzidien et al. 2022; Razon et al. 2022; Salaün et al. 2022; Hemmat et al. 2023; Abeywardena 2023; Mohammadi et al. 2024).

Topic modelling, particularly through LDA, has become a prevalent method in the analysis of legal texts across various jurisdictions. This technique has been utilised to model case law from a range of courts, including the Australian High Court (Carter et al. 2016), Supreme Court of the Netherlands (Remmits 2017), Brazillian Supreme Court (Luz De Araujo and De Campos 2020), Philippine Supreme Court (Razon et al. 2022), Canadian Housing Tribunal (Salaün et al. 2022), UK contract law cases (Izzidien et al. 2022), the UK Supreme Court and Sri Lankan Supreme Court (Abeywardena 2023) and the European Court of Human Rights (Mohammadi et al. 2024).

Luz De Araujo and De Campos (2020) performed an LDA on 45,532 cases that were manually annotated by the Brazillian Supreme Court experts. The paper compared LDA topic-based representations with traditional bag-of-words representations and demonstrated weighted F1 scores of 78.07 for the LDA, 89.22 for the TF-IDF, and 89.29 for word counts. The accuracy of LDA in representing legal cases was further evidenced by Razon et al. (2022), who found that the LDA topics match the cases they describe with 86.67% accuracy. Salaün et al. (2022) expanded their analysis to include LDA, Latent Semantic Indexing (LSI) and BERTopic models. Following the BM25 proximity score, the LDA was the highest performing, with an F1 score of 53.2. By the SBERT cosine similarity metric, BERTopic achieved the highest score of F1 57.4. Their study also highlighted the difficulties in evaluating topics due to the necessity for manual classification prior to modelling. Manual classification may also vary by relevant legal and domain expertise (Remmits 2017).

While these studies demonstrate the utility of LDA in legal contexts, they also reveal limitations. Carter et al. (2016) observed the challenge of topic modelling arising from the difference between operational definitions within case law, which often diverge from theoretical and traditional understandings of legal areas. Further, Izzidien et al. (2022) discussed the challenges posed by the granularity of legal issues, which can lead to the generation of various subtopics, complicating the model's application. Similarly, Abeywardena (2023) needed to use multiple topics with often overlapping labels to accurately classify the LDA topics. Mohammadi et al. (2024) used an LDA model to analyse ECHR Article 8 decisions and judgments relating to eviction (identified by existing keywords), selecting the 17 most prominent topics based on the model's coherence scores and human judgment. The authors elaborated on these limitations: "LDA did not reveal more specific topics, such as evictions. Identifying these more specific subtopics would, however, be very useful for practitioners and researchers" (Mohammadi et al. 2024). Dyevre (2021a) uses eight computer-aided text-scaling approaches for the analysis of jurisprudential change and opinions. The paper reveals two challenges for judicial opinion mining; the multidimensionality and complexity of judicial texts and the challenges of producing reliable human annotations of long, complex, jargon-heavy judicial opinions making validation difficult. Dyevre (2021b) also presents a structured overview of these techniques and how it may reveal patterns and trends in large legal corpora, enabling more nuanced interpretations of judicial behaviour and legislative evolution.

LDA topic modelling has also been applied to other legal texts, including UK legislation (O'Neill 2016), Latvian legal documents (Vīksna et al. 2020), French Parliamentary Debates (Bourgeois et al. 2022) and Iranian laws and parliamentary documents (Hemmat et al. 2023). Soh et al. (2019) evaluated various text classification techniques using a pre-labeled corpus of case law. Among the models tested, a linear support vector machine trained on topics extracted via LSA demonstrated a macro F1 score of 63.2 and a micro F1 score of 73.3. In Soh Tsin Howe (2024), BERTopic, alongside LSA, was employed to generate topics for a dataset encompassing domain name disputes and a canonical collection of cases from the European Court of Human Rights. LSA performed better, with the highest MDR (43.1%) and AUROC (91.4%). Aguiar et al. (2022) use BERTopic to build thematic models across Brazillian Supreme Court cases and relevant legislation. The paper shows a macro F1 score of 89.0, although it only categorises documents across five classes of the Brazil National Council of Justice: Common Civil Procedure, Execution of Extrajudicial Title, Criminal Action – Ordinary Procedure, Special Civil Court Procedure, Tax Enforcement; and the option for Other (which makes up 21.0% of the dataset). BERTopic was also used on 314 cases from the Cornell Legal Information Institute's Historic US Supreme Court Decisions repository. It achieved a semantic correspondence of 84.6% between the topics generated and the documents they represent, as validated by expert evaluations (Silveira et al. 2021).

A critical component in the application of traditional topic modelling methods is the role of human annotators and human judgment. The process of topic allocation in these methods typically relies on the interpretation of a "bag-of-words" or vector representations derived from the text data. This exploratory process often results in a heterogeneous mixture of topic abstractions, where some topics are finely detailed while others may overlap, thereby compromising the orthogonality between topics. Further, the lack of agreement on how to choose the best models and evaluate results in topic modelling makes this analysis challenging (Rüdiger et al. 2022).

In the study by Wendel et al. (2022) on the German Federal Constitutional Court, the authors generated a topic model of decisions published from 1951 to 2017. To mitigate bias in topic interpretation, they consulted experts familiar with the court's decisions using a catalogue of questions to predict expected topics. The variability in experts' responses underscored the ambiguous nature of what precisely constitutes a topic. The study employed structural topic modelling as an initial step to define these topic terms and developed an 'area of law' to categorise and select specific legal areas. However, the overlapping and hierarchical nature of these areas underscored the complexities involved in creating a definitive taxonomy of law, leading to the conclusion that developing a comprehensive legal taxonomy was neither feasible nor necessary for their analytical purposes.

The observed limitations, including the variable accuracy of traditional topic modelling, if measured at all, and the prevalence of general and overlapping topics, have led us to pursue a new method. Traditional models like LDA, LSA and BERTopic have shown constraints in handling the nuanced complexity of legal texts. Traditional approaches to topic modelling also involve meticulous parameter tuning—selecting the number of topics and adjusting alpha and beta values—which can be both time and computationally intensive and susceptible to subjective biases in interpreting the outputs (Rüdiger et al., 2022). In contrast, LLMs offer the potential for instantaneous topic generation, which bypasses some of these challenges. Mu et al. (2024a) explain that the transformative nature of LLMs increases the model's "ability to comprehend context, nuances, and even subtle thematic undertones has been demonstrated in various NLP tasks, allowing for a richer and more detailed categorisation of topics". Dyevre (2021b) advocates for using machine learning as a supplement to human analysis rather than a replacement, noting that legal language's complexity requires continuous human interpretation, especially in unsupervised tasks. This perspective aligns with recent trends in legal NLP, where ML models like transformers have shown promise in parsing complex legal discourse while recognising the limitations of fully automated analysis.

While we have not identified any paper using LLMs to comprehensively topic model case law, recent papers have engaged with using LLMs instead of traditional NLP methods in areas outside of law.

A notable advancement in the field is presented by Wang et al. (2023), introducing PromptTopic, a three-stage topic modelling method that leverages the capabilities of LLMs to refine and categorise topics. The paper demonstrated comparable performance to established methods such as LDA, NMF and BERTopic, and even surpassed them in generating more meaningful topics upon qualitative assessments (Wang et al. 2023). Xu et al. (2023) propose a novel framework named Diffusion-Enhanced Topic Modeling using Encoder-Decoder-based LLMs (DeTiME). DeTiME uses Encoder-Decoder-based LLMs to produce improved clustering of embeddings to generate topics that exhibit enhanced semantic coherence.

One similar paper to our work by Drápal et al. (2023) proposed a framework using an LLM to perform thematic analysis of fact summaries from criminal court opinions regarding theft. The method involved collaboration between human legal experts and Chat-GPT4 to generate initial theme codes, search for themes and classify the data. Unlike our proposed method, the authors do not use the full text of case law nor do they propose a defined list of topics for the LLM to select from. The shift towards LLMs in legal topic classification represents a significant evolution in computational legal studies.

## 3 Functional Legal Taxonomy

Understanding the topics addressed in court is important from multiple perspectives. For policymakers, recognising the areas most frequently litigated helps identify social conflicts and the role courts play in resolving them (OECD 2024). For court administrators, insights into common legal topics assist in resource and capacity planning. Additionally, for those using AI to analyse court decisions, knowledge of prevalent legal topics aids in identifying gaps in training datasets and enhancing the predictive accuracy of relevant models. Assigning a topic to each case can improve the performance of classification models, particularly when predicting case outcomes. For example, court decisions from the same legal area can be used in few-shot prompting to improve the results of LLM-based judgment prediction (Xie et al 2024). Using an LLM to extract principal legal features from a case allows for the automatic identification of the case's topic from a predefined list. By maintaining consistency through the predefined list of topics, the reliability of the classification is reinforced.

Currently, there is no standardised taxonomy of legal areas suitable for widespread application in topic or legal area classification within the UK. Existing taxonomies, including the Standards Advancement for the Legal Industry (**SALI**) Taxonomy (SALI Alliance, n.d.), the Stanford Legal Design Lab's Legal Issues Taxonomy (**LIST**) (Stanford Legal Design Lab, n.d.), theoretical work on legal taxonomies (Sherwin 2009) and qualitative approaches to empirical legal research (Webley 2010) have informed the principles and details of our approach. However, existing taxonomies are often tailored to specific jurisdictions or legal fields and are, thus, not usefully applicable to UK law due to their narrow scope and regional focus. This applies, in particular, to the SALI and LIST taxonomies. To give just one example, SALI has a US focus and uses the US term 'bankruptcy' instead of the UK term 'insolvency'. This is relevant, since in UK terminology 'bankruptcy' and 'insolvency' refer to different procedures. As defined in the UK Insolvency Act 1986, the former only applies to natural persons while the latter concerns companies. This is not to question the usefulness of existing taxonomies such as SALI and LIST. These are very helpful initiatives. However, their jurisdictional focus makes them less relevant for the task at hand in this paper, which has a particular focus on UK law.

To address this gap, we have developed a comprehensive taxonomy tailored to the UK legal system. We would have preferred to avoid this and use an existing UK taxonomy. However, our extensive research showed that up to this point, there was no universally accepted taxonomy among UK legal institutions. Our search covered, for example, publications and websites by ministries, courts, universities and professional bodies like the Law Society or the Solicitors Regulation Authority (**SRA**). Our review showed that no standardised classifications were available from educational institutions or for professional examinations like the Solicitors Qualifying Examination. Against this background, we offer a new taxonomy of legal areas taking a UK perspective.

Our taxonomy adopts a functional approach, focusing on practical legal issues faced by citizens and businesses. A functional approach focuses on the effects of law in reality (Michaels 2013), thereby ensuring that the classifications reflect the interests and concerns of those concerned with the legal matter. Functional, as opposed to doctrinal, approaches capture what is relevant both for individuals and for their relationship to society and its institutions (Steffek 2013). Functional taxonomies also come with the advantage of facilitating cross-jurisdictional comparisons since they provide a yardstick external to local laws. Applying a functional methodology, we strive for clarity and differentiation, ensuring that the categories of our taxonomy are distinct and comprehensible. As a starting point, we used The Law Society's list of areas of practice provided in its 'Find a Solicitor' service (The Law Society, n.d.). This service allows users to find solicitors based on their areas of practice. This fits well with our functional framework since we are aiming for distinguishable areas of law from the perspective of real-world problems. However, this list needed to be amended as its focus is the practice of solicitors, while our aim is a taxonomy of legal areas for the entire UK legal system. Appendix B documents the 44 amendments we have made to The Law Society's areas of practice, resulting in a new and original taxonomy of 108 specific areas of law. In the amendment process we applied the grounded theory method (Webley 2010). Grounded theory refers to approaches where the theory is developed as the research proceeds. In our case, this meant that we amended the taxonomy by adding, changing and deleting categories to better reflect the cases in our dataset. For example, we added the category of 'Health' to reflect the relevant number of cases in the dataset concerning health matters. The term 'Health' instead of 'Medical negligence' or similar also illustrates the functional approach, i.e., an approach that aims to capture real world phenomena as opposed to legal doctrine. We also streamlined the taxonomy by simplifying it where necessary. For example, we deleted ‘Immigration – asylum – legal aid (IML)’ and amended ‘Immigration – general – legal aid (IMG)’ to ‘Immigration – legal aid (IML)’ with the effect that the latter now captures the two former categories.

As a result, our taxonomy uses modern functional language bridging law and disputes in reality. Such a taxonomy is useful for both lawyers and laypersons, as the language used will be understood by both groups (cf. Salaün et al. 2022). Given its focus on modern law and reality, it is particularly well suited to represent more modern legal texts. The taxonomy might need to be adapted for legal documents based on language that is materially different from modern language, e.g., Medieval legal texts (cf. Grajzl and Murrell 2022).

The taxonomy ranges from ‘Administrative and public law (PUB)’ to ‘Landlord and tenant – leasehold valuation (LLV)’ and ‘Travel and tourism (TAT)’. Each area is assigned a three-letter abbreviation for ease of reference. Where necessary, legal areas are further subdivided to capture more specific case types, such as various aspects of landlord and tenant law. Based on the grounded theory approach, we provide differentiations as required for legal areas that have a higher number of cases. Generally, each area that is further differentiated contains a general category, which allows for capturing cases that cannot clearly be assigned to one of the more specific areas. The ‘Landlord and tenant’ theme, for example, is further divided as follows:

- Landlord and tenant – general (LRG)
- Landlord and tenant – commercial (LCO)
- Landlord and tenant – leasehold valuation (LLV)
- Landlord and tenant – residential (LRE)

The specific areas of law are placed into one of the following overarching higher-level areas of law:

- Public law
- Dispute resolution law
- Commercial law
- Personal and consumer matters
- Criminal law
- International law

The full taxonomy is shown below in a radial plot (Figure 1).

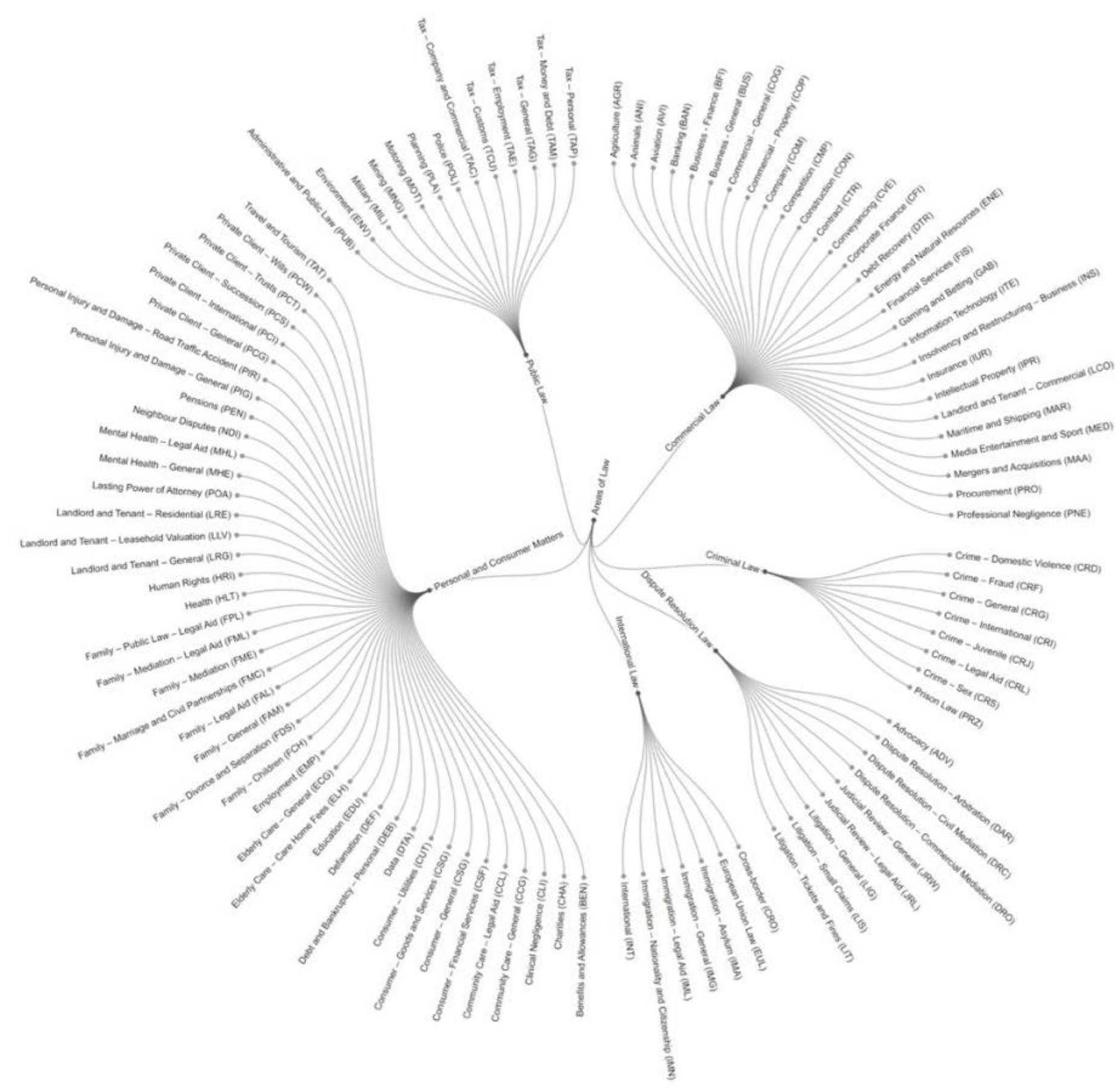

*Figure 1. Radial plot of the six higher-level areas of law, with the respective specific areas of law*

## 4 **Data and Methodology**

### 4.1 *Cambridge Law Corpus: Summary Judgment Cases*

The data used for this study consists of UK case law from the Cambridge Law Corpus (Östling et al. 2023).[2] While the Cambridge Law Corpus contains around 320,000 cases, we specifically use a subset containing only summary judgment cases that was published by Izzidien et al. (2024). The dataset contains 3,078 summary judgment cases across 30 courts and tribunals in the UK. Each case is held in extensible markup language (**XML**), including a set of tags stored in a Python pandas data frame. The tags include the court name, last date of the hearing's delivery, neutral citation, word count, court tier and the full text of the case.

While summary judgment procedures have existed since the 1800s, the modern procedure was redefined in the *Civil Procedure Rules 1998*, which entered into force on 26 April 1999.[3] Therefore, the subset of summary judgment cases are only cases from 26 April 1999 to 7 June 2023.

[2] For access to this dataset see: https://www.cst.cam.ac.uk/research/srg/projects/law.

[3] Civil Procedure Rules 1998, pt 24 (in force 26 April 1999); made under authority of the Civil Procedure Act 1997.

## 4.2 *Large Language Model: Claude 3 Opus*

We use Claude 3 Opus, an LLM, to classify each summary judgment case with a topic from the functional legal taxonomy. There are two primary reasons for selecting Claude 3 Opus. According to Anthropic's model card for Claude 3 Opus, it demonstrates superior performance across several benchmarks compared to its predecessors and contemporaries (specifically GPT and Gemini) (Anthropic 2024a; Anthropic 2024b). In addition, it has a 200,000 token context window, unlike GPT's 128,000 token context window, allowing us to work more easily with longer court judgments (Anthropic 2024a; Anthropic 2024b). The primary reason for selecting Claude 3 Opus is the token context window. Of the 3,078 cases, 1,100 cases are above 8,000 words, meaning we required the relatively longer 200,000 token context window.

In a detailed comparison of Claude, GPT and Gemini models, it was found that Claude 3 Opus consistently outperforms the others across a spectrum of tasks, particularly those that test knowledge and reasoning abilities—key components in legal analysis. In benchmarks focused on general reasoning, Claude 3 Opus achieved 86.8%, compared to GPT-4 at 86.4%, GPT-3.5 at 70.0%, Gemini 1.0 Ultra at 83.7% and Gemini 1.0 Pro at 71.8%. Additionally, on benchmarks for reading comprehension, Claude 3 Opus achieved a score of 83.1%, compared to GPT-4 at 80.9% and Gemini 1.0 Ultra at 82.4%. Claude 3 Opus has also demonstrated strong performance on mixed evaluations and knowledge Q&A (Anthropic 2024b). Claude Opus surpassing ChatGPT-4 on the Alice in Wonderland reasoning problem (Nezhurina et al 2024) and, according to a study by Kevian et al. (2024), Claude 3 Opus not only excelled in general cognitive tasks but also demonstrated superior performance to Gemini Ultra and Chat GPT-4 in specialised tasks.

## 4.3 *Prompt development for topic classification*

The development of an effective prompt for Claude 3 Opus in the classification of legal topics from summary judgment cases was a key step in our methodology. Drawing on recent advancements in prompt engineering and the self-referential capabilities of LLMs, our approach was guided by the prompt design elements set out here and further refined through our preliminary applications.

Initially, we tasked Claude 3 Opus with the creation of a custom prompt designed to guide the classification of summary judgment cases. The prompt instructed the model to allocate each case to a specific legal topic from a provided list, requiring it to identify both primary and potentially secondary topics relevant to the case. The prompt was designed to include several key components:

1. Task framing and context setting: The prompt begins by clearly framing the task, instructing Claude 3 Opus to review the case details and assign the most fitting legal topics based on the functional legal taxonomy provided within the prompt.
2. Closed-set prompting: By using a closed-set prompting strategy, we constrained the model's responses to a predefined list of topics, enhancing the focus and accuracy of the output (White et al. 2023; Mu et al. 2024).
3. Instruction prompting: The prompt provides detailed instructions that guide the model in its task. The prompt also asks that the response meet the criteria and emphasises the importance of attention to detail (Chang et al. 2024).
4. Examples and constraints: The prompt explicitly listed the legal topics as selectable options, serving both as constraints and examples to aid the LLM in understanding the expected response format, adhering to the guidance from Anthropic's prompting guide (Anthropic 2023).
5. Encouraging reasoning and explanation: The LLM was prompted to provide reasoning and explanations for its topic selections, enhancing the transparency and traceability of its decisions. However, the level to which these forms of explanations correspond to the actual steps taken by the model remains an open question (Paul et al. 2024; Agarwal et al. 2024).

6. Contextual prompting: Additional contextual information relevant to topic modelling was incorporated to guide the LLM toward the desired output (Liu et al. 2021).
7. Few-shot prompting: The setup subtly introduced a few-shot learning approach by including a list of topics from which the LLM must choose Agrawal et al. (2022).

Following the initial prompt developed by Claude 3 Opus, feedback was solicited directly from the LLM, leading to additional refinements. Below are five suggestions provided by Claude 3 Opus that were incorporated into the final prompt.

1. Clarification of output format: Explaining that the number assigned to the topics was not part of the output format should help avoid confusion.
2. Emphasis on relevance: Specifying that the selected topics should be the most central to the case and that peripheral issues should not influence the topic allocation.
5. Request a confidence score: Including a request for a confidence score (e.g., on a scale of 1-5) for its topic allocation could help assess the reliability of its decision. This method, in principle, builds on earlier work in neural networks (Hendrycks & Gimpel, 2018; Li et al. 2024a). However, since LLMs generate their output sequence in an auto-regressive way, they do not directly produce a confidence score for the output sequence (Chen et al. 2023).
3. Introduction of an example: Defining a sample case and its corresponding topic allocation would be useful to serve as a reference point for Claude Opus.
6. Specification of response length: Indicating the desired length of Claude Opus's explanation for its topic selection should ensure a concise yet informative response.

The full prompt and exchange are in Appendix A.

#### *4.3.1 Initial legal expert checks on topic results*

In our study, we conducted an initial evaluation of Claude 3 Opus's ability to classify the topics using a sample of 50 cases. This assessment aimed to identify the accuracy of the topic classifications made by the LLM and to pinpoint specific areas where the model's performance could be improved.[4]

The initial manual check revealed that 35 out of 50 cases (70%) were classified correctly. However, 15 cases (30%) presented issues with the classifications. First, three cases were incorrectly classified. Second, in several cases where the model had identified multiple relevant topics, the order of prioritisation of these topics or one topic being correct while another incorrect was identified as an issue. Therefore, the prompt was amended to remove the option for multiple topics. Third, we identified two cases that had used similar wording or abbreviations to the list of topics but were slightly different. These deviations can be defined as "hallucinations", a critical problem for natural language generation by LLMs, where the generated content is "unfaithful to the provided source content" (Ji et al. 2023; Farquhar et al. 2024). To reduce instances of such hallucinations of the topic name, we incorporated the following additional instructions:

> "Before providing your response, please ensure that the selected primary and secondary topics are actually present in the list of legal topics provided above. If the topics you have chosen are not found in the list, please reconsider your selection and choose topics that are included in the list that meet the description of the primary and secondary topic."

This modification mitigated the error rate and underscored the importance of iterative prompt refinement in the effective use of LLMs for legal topic classification.

---

[4] The manual checks were conducted by an author of this paper, a qualified solicitor with higher degrees in law.

### 4.3.2 *Statistically significant legal expert checks on topic results*

We then ran the final prompt across all cases, producing a classification from Claude 3 Opus. We then conducted manual checks on a sample. In instances where the classification was incorrect, the annotator recorded the correct label. To ensure a statistically significant sample, we calculated a sample size of 342 cases for a 95% confidence level with a 5% margin of error for the 3,078 cases. We excluded these 50 cases used to optimise the prompt from the statistically significant sample. The sample was randomly drawn from the dataset utilising Python's random number generator.

## 5 **Results**

### 5.1 *Legal expert evaluation*

A comprehensive evaluation was conducted using a statistically representative sample of 342 cases to assess the performance of Claude 3 Opus in classifying legal topics. The legal expert evaluation revealed that Claude misclassified 44 cases, yielding an initial accuracy score of 87.13%.

However, it was clear from the review of the cases and the assigned topics that several of these misclassifications involved only minor discrepancies in topic naming, while the general category was otherwise correct. For example, Claude 3 Opus classified a case as "Dispute resolution - adjudication (DAR)" when it should have followed the correct language of the taxonomy being "Dispute resolution – arbitration (DAR)" or should have classified it by the correct term "Public procurement (PRO)", but Claude had classified as just "Procurement (PRO)". There were eight minor discrepancies and of these eight four were discrepancies where the three-letter abbreviation matched the taxonomy correctly, even though the complete name did not fully align. Adjusting only for these four minor errors where only naming variations occurred and the abbreviation was correct, the accuracy score improves to 88.20%. We also calculate accuracy itemised by legal area in Appendix D. These results provide more granular information to contextualise the visualised results below that focus on different legal areas.

Three additional types of error were identified in the legal expert evaluation. First, there were 26 cases where the model had erroneously assigned primary topics that were incorrect while the secondary topics were correct. Common errors included where the secondary topic was "Contract (CTR)", but Claude had incorrectly identified the primary topic as "Dispute resolution – civil mediation (DRC)". Notably, 25 of these cases arose when the primary topic was one of the dispute resolution categories, and only one was a case classified as "Company (COM)" when it should have been the secondary topic "Insolvency and restructuring – business (INS)". Such results suggest that the general nature of dispute resolution categories was unclear to the model.

Second, the model generated topics for 4 cases that were not part of the legal taxonomy provided. Each of these "hallucinated" topics is practically and contextually correct; they each were relevant and similar to topics from the taxonomy:

1. Classified as a hallucinated topic "Confidentiality (CON)". The correct topic from the taxonomy would be "Contract (CTR)" as it deals with a confidentiality agreement.
2. Classified as a hallucinated topic "Partnerships (PAR)". The correct topic from the taxonomy would be "Company (COM)" as it deals with issues of partnership liability in a company law context.
3. Classified as a hallucinated topic "Restitution (RES)". The correct topic from the taxonomy would be "Contract (CTR)" as restitution is a contract law remedy.

4. Classified as a hallucinated topic "Costs (COS)". The correct topic from the taxonomy would be "Litigation – general (LIG)" as costs are an element of general litigation.

Finally, 14 cases were incorrectly classified. Of these cases, the incorrect classifications were one of four topics: "Dispute resolution – civil mediation (DRC)" (6 misclassifications); "Dispute resolution – commercial mediation (DRO)" (2 misclassification); "Landlord and tenant – general (LRG)" (3 misclassifications), and "Landlord and tenant – residential (LRE)" (3 misclassifications). Similar to the above issues of primary and secondary topics, it appears these specific areas of law caused confusion in the model regarding their general nature.

### 5.2 *Additional legal expert evaluations for topic hallucinations*

Given the results on minor discrepancies in topic naming and hallucinations on topics, we compared the Claude topic output across the whole dataset with the taxonomy. We identified, in addition to the original 13 hallucinations identified above, 45 of the 3078 cases, which had also been classified with a topic not contained in the taxonomy (both minor and major hallucinations). This confirms that 98.28% of the Claude responses used the topics from the taxonomy.

In these secondary evaluations of these errors, similar issues arose, as described above. There were 5 cases that had the correct topic classification but were missing the relevant abbreviation, i.e., "Banking" instead of "Banking (BAN)". Additionally, 29 cases had minor hallucinations that were materially correct but had some semantic errors, i.e., "Fraud – crime (CRF)" instead of "Crime – Fraud (CRF), or "Trusts (PCT)" instead of "Private client – trusts (PCT)". Finally, there were 14 cases with major hallucinations. Similar to the above evaluation, these hallucinations were contextually correct but often hallucinating a more specific topic than was provided for by the taxonomy. For instance, cases classified as "Carriage of goods by sea (COG)" were corrected to "Commercial – general (COG)". Similarly, cases classified as "Civil procedure (CIP)", "Contempt of court (CON)", "Costs (COS)", and "Limitation (LIM)" were corrected to "Litigation – general (LIG)". Cases classified as "Confidential information (CFD)" and "Restitution (RES)" were corrected to "Contract (CTR)", while those classified as "Confidential information (CON)" were corrected to either "Intellectual property (IPR)" or "Defamation (DEF)". Additionally, "Conflict of laws (COL)" was corrected to "Cross-border (CRO)", "Crime – blackmail (CRB)" was corrected to "Crime – general (CRG)", "Partnership (PRT)" was corrected to "Company (COM)", "Privacy (PRI)" was corrected to "Human rights (HRI)", and "Regulation - legal profession (REG)" was corrected to "Professional negligence (PNE)". These hallucinated topics can be understood as narrower constructions of specific legal issues under the broader topic outlined in our taxonomy. While not in the scope of this current research, we look forward to further work that considers why such hallucinations arise and how they can be avoided. We have manually corrected the topic for these 58 instances.

### 5.3 *Metric results*

To assess Claude's classification performance, we compared its predictions against the correct labels identified by the expert annotator and calculated several standard evaluation metrics: accuracy, precision, recall and F1 score. These were computed both at the overall dataset level and on a per-class basis. In addition, we derived macro F1, micro F1 and weighted F1 scores to gain a more nuanced understanding of the model's strengths and weaknesses across different classes.

While accuracy, precision, recall and F1 scores represent standard evaluation metrics within the field of machine learning, we briefly restate these concepts for clarity. Accuracy, in this context, refers to the proportion of correctly classified instances among the total number of instances under review. Precision expresses the proportion of instances classified as positive that are truly positive, thereby reflecting the extent to which the model avoids mislabelling negative instances. Recall identifies the proportion of truly positive instances that the model successfully classifies as positive, thus demonstrating the model's sensitivity or ability to detect the relevant phenomenon. The F1 score, which is the harmonic mean of precision and recall, strikes a balance between these two metrics and provides a single measure that encapsulates overall performance. Macro F1, micro F1, and weighted F1 scores offer more nuanced perspectives on these values by taking into account variations in class size and class imbalance, ensuring that no single category disproportionately influences the evaluation results.

Table 1 sets forth the overall metrics, summarising the classifier's performance across all classes.

*Table 1. Overall Classification Metrics*

| **Metric** | **Value** |
|---|---|
| Accuracy | 0.87 |
| Weighted Precision | 0.94 |
| Weighted Recall | 0.87 |
| Weighted F1 | 0.89 |
| Macro F1 | 0.87 |
| Micro F1 | 0.87 |

The balanced macro and micro F1 scores, when considered in conjunction with the weighted F1, suggest that the classifier manages both large and small classes reasonably well. Nevertheless, the presence of certain classes with lower precision, recall or F1 values warrants further examination. These classes are illustrated in Appendix D, where the accuracy, precision, recall and F1 scores are reported for each category, accompanied by the support, which indicates the number of ground truth instances available for that label. Larger supports instil greater confidence in the reliability of the reported metrics, whereas smaller supports may produce inflated or unstable scores.

As demonstrated in Appendix D, different performance is achieved across different areas of law; several categories achieved relatively lower F1-scores. Table 2 includes the legal areas with an F1-score below 0.80.

*Table 2. Classification metrics for lowest scoring areas of law*

| **Area of Law** | **Precision** | **Recall** | **F1-score** | **Support** |
|---|---|---|---|---|
| Crime - fraud (CRF) | 1.00 | 0.33 | 0.50 | 3 |
| Dispute resolution - civil mediation (DRC) | 0.16 | 1.00 | 0.27 | 3 |

| Dispute resolution - commercial mediation (DRO) | 0.08 | 1.00 | 0.14 | 1 |
|---|---|---|---|---|
| Human rights (HRI) | 0.00 | 0.00 | 0.00 | 1 |
| Landlord and tenant - general (LRG) | 0.40 | 1.00 | 0.57 | 2 |
| Planning (PLA) | 0.00 | 0.00 | 0.00 | 6 |

For example, Planning (PLA), despite having a support of six instances, yields an F1 score of zero. Similarly, certain Dispute Resolution subcategories, including Civil Mediation (DRC) and Commercial Mediation (DRO), exhibit poor performance, each with a support of three. Closer scrutiny of these underperforming topics reveals that certain metrics, such as precision or recall, may appear deceptively robust when drawn from a minimal number of test instances. For instance, while Crime – fraud (CRF) attains a precision of 1.00 on a sample of three instances, its recall is limited to 0.33. Extrapolating this result to a larger sample of 47 known CRF cases (see Figure 4) within the broader dataset indicates that the model would likely miss the majority of these cases.

The implications of these findings are twofold. First, at a practical level, the identified inconsistencies suggest that the classifier's reliability may vary significantly depending on the nature and frequency of the legal issues at hand. Categories with lower F1-scores may necessitate closer monitoring, manual verification or additional training data to ensure accurate classification. Second, from a methodological standpoint, these results highlight the importance of critically interpreting performance metrics in light of the underlying data distribution and sample size. Legal domains that are underrepresented in the training set, or that encompass more nuanced or context-specific legal concepts, may require targeted refinement of the classification approach. Ultimately, these observations reinforce that, while automated classification systems can offer valuable initial insights, their outputs must be subjected to ongoing evaluation and, where necessary, complemented by expert legal judgment.

Lastly, the classification for the original dataset of SJ cases provided by Izzidien et al. (2024) had a weighted F1-score of 0.94. This introduces a cascading error effect through subsequent classification (weighted F1 of 0.89). These combined errors imply that, on average, 16.34% of predictions may be impacted by either the dataset's inherent inaccuracies or classification errors – assuming error independence. With compounded uncertainties, the reported 87.13% accuracy should thus be interpreted as an upper bound. Given that this represents multi-label classification of legal case topics, with not all cases being summary judgment cases, the support for each individual label should be interpreted with additional caution. The true relaibility in any given topic label is likely lower than the aggregate metrics suggest, as errors may distribute unevenly across different topic categories and case types. Additional care in interpreting results where there is low support may be needed.

Outside these problematic categories, the remaining classes are generally identified with considerable robustness. However, small support values remain a limitation in many instances, restricting the extent to which one can confidently generalise the reported metrics.

### 5.4 *Visualised results*

Once all manual checks were completed and hallucinations corrected, we used the final primary topic classifications to visualise the topic distribution across the summary judgment cases. We used the name of the court for each case, the date of hearing and the court tier list in Appendix 3 for our analysis. The following visualisations aid in understanding the trends and patterns in summary judgment cases. A detailed discussion of these results can be found in Section 6.1.

### 5.4.1 *Proportions of topics within higher-level areas of law*

First, we plot the number of summary judgment cases per higher-level areas of law. Figure 2 shows that commercial law is the dominant higher-level area of law, followed by dispute resolution, then personal and consumer matters. Criminal, public and international law have the fewest summary judgment cases.

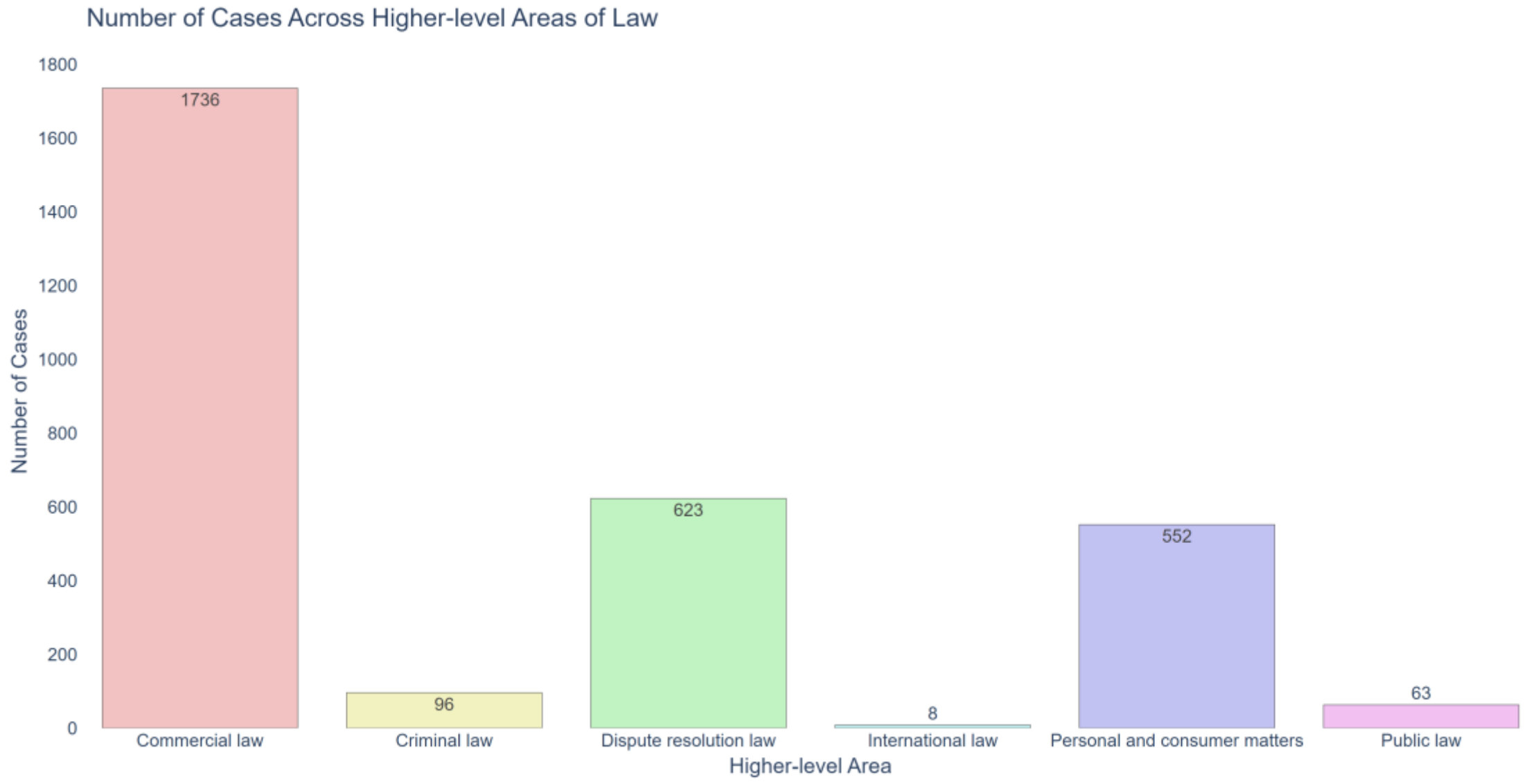

*Figure 2. Number of summary judgment cases per higher-level area of law*

The following Figures 3 to 8 show the numbers of summary judgment cases per specific area of law grouped by the five higher-level areas. These visual representations provide a granular view of the distribution of cases, highlighting the prevalence and thematic concentration within each higher-level area of law. This is key to understanding the judicial application of summary judgments in different contexts.

Figure 3 illustrates the breakdown of specific areas of law within the higher-level area of commercial law. It shows the highest concentration of cases in contract, followed by intellectual property, construction, company, banking, and professional negligence.

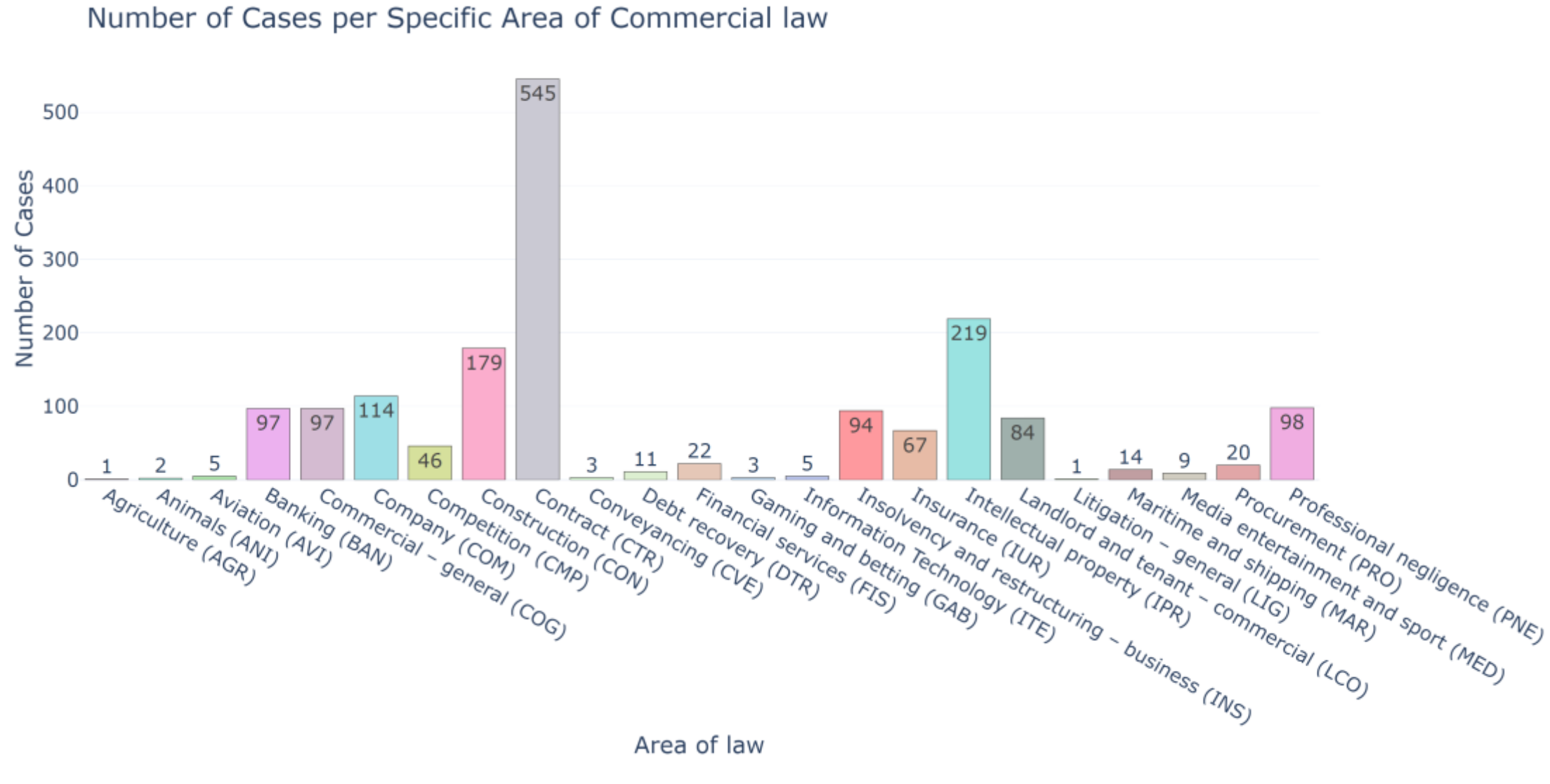

*Figure 3. Number of summary judgment cases per specific area of commercial law*

Summary judgment cannot be used in actual criminal cases; however, it can be used in civil proceedings following a criminal case. For example, a civil case running alongside a claim of fraud or a suit for damages for wrongful arrest. Figure 4 reveals primarily cases relating to fraud and general criminal topics. These are the most likely areas for summary judgment to overlap with a criminal law topic.

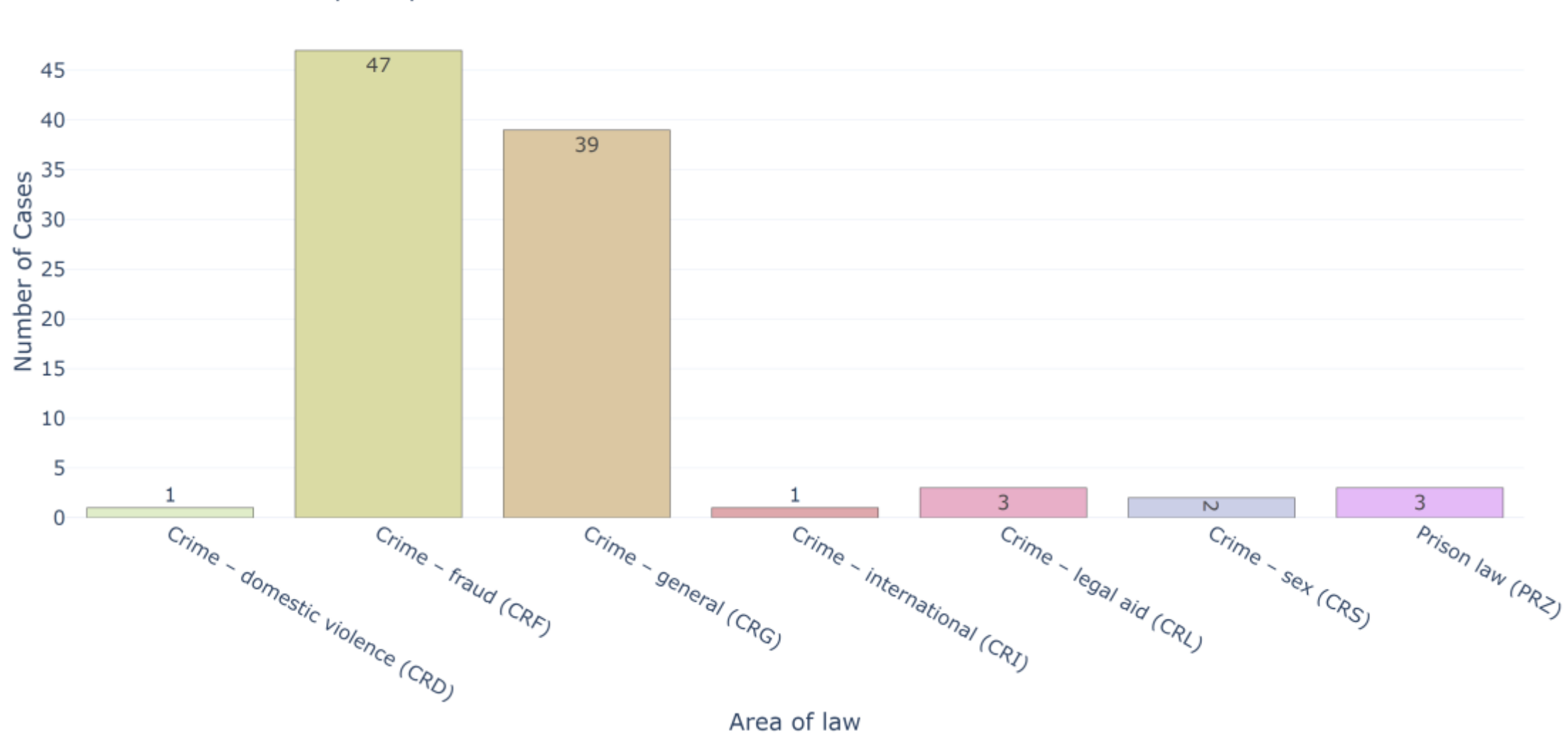

*Figure 4. Number of summary judgment cases per specific area of criminal law*

Figure 5 focuses on dispute resolution, highlighting arbitration, civil mediation and general litigation as the dominant topic areas. The data suggests the utilisation of summary judgments in scenarios where arbitration or mediation processes are either concluded or not feasible, prompting a judicial decision based on the available facts. This underscores the role of courts in affirming outcomes in alternative dispute resolution mechanisms or in deciding cases that may not require a detailed examination of complex evidentiary matters. It also shows its use in general litigation matters to resolve administrative matters such as determining costs, affirming judicial awards or remedies, and determining procedural issues.

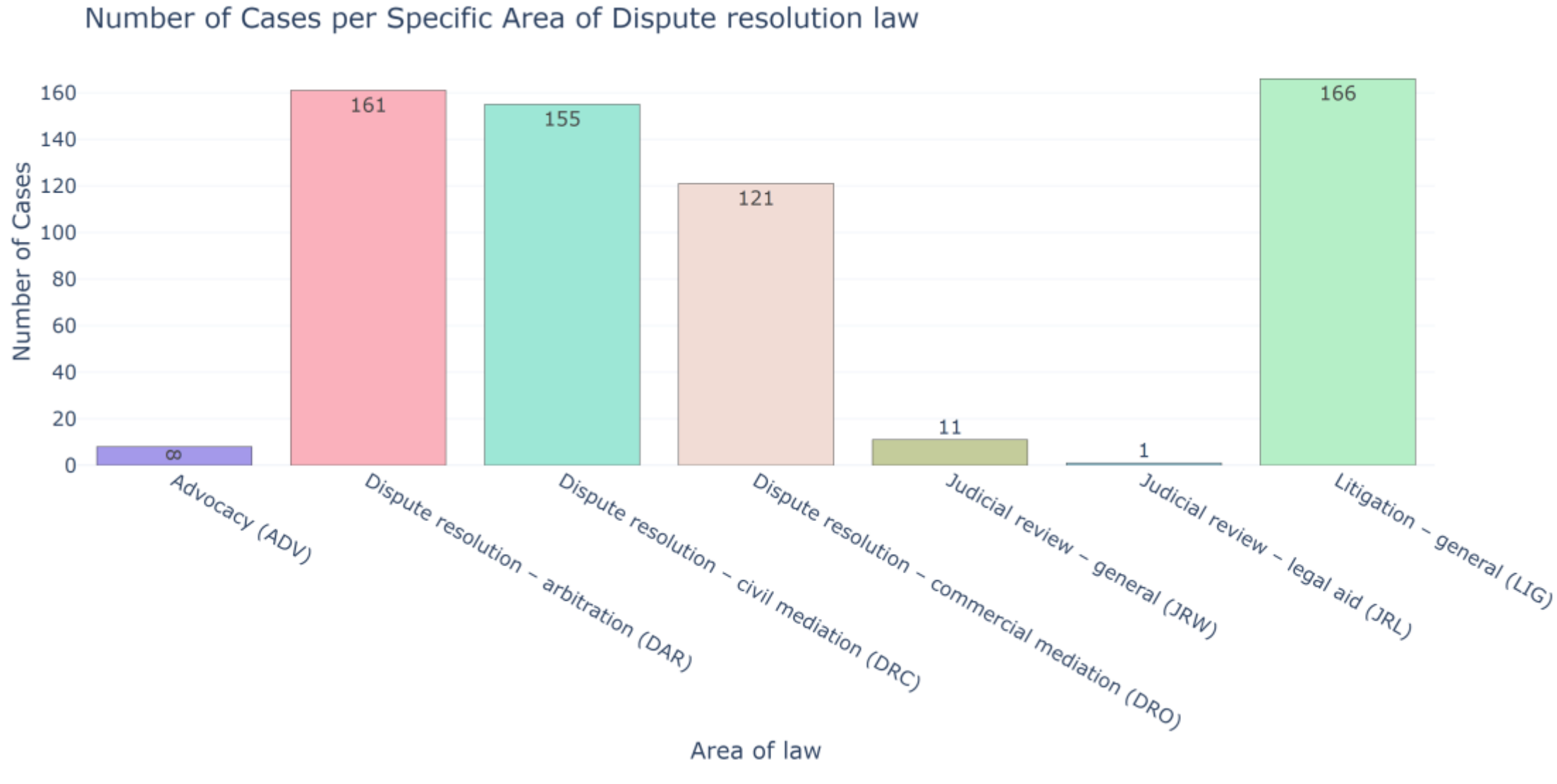

*Figure 5. Number of summary judgment cases per specific area of dispute resolution law*

Figure 6 shows the use of summary judgment in international law cases, with few cases overall but primarily focused on European Union law and international law. Summary judgments in these cases might occur when legal questions are clear and do not require extensive deliberation over jurisdiction or aspects of foreign laws.

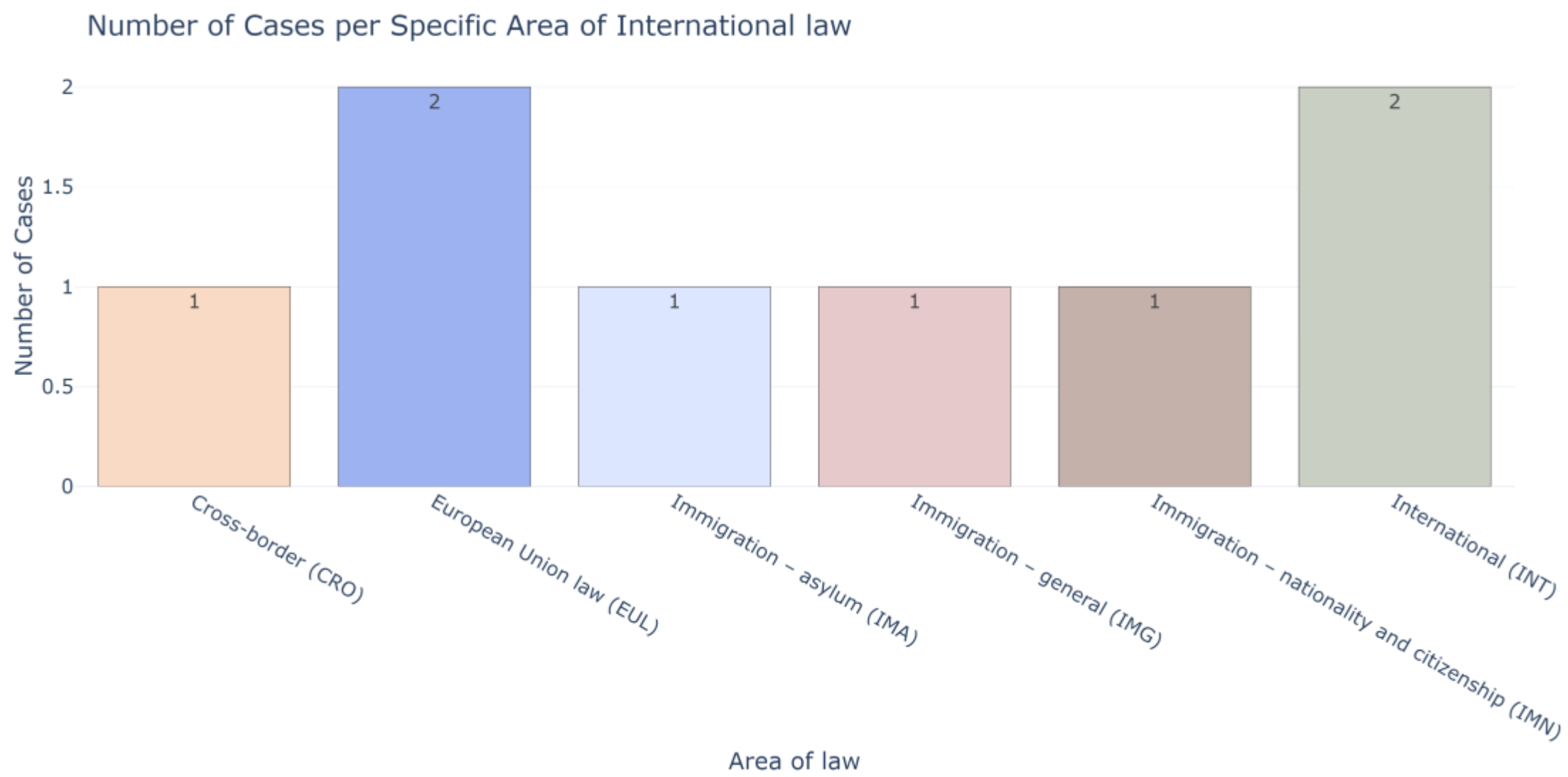

*Figure 6. Number of summary judgment cases per specific area of international law*

Figure 7 illustrates the breakdown of summary judgment cases within personal and consumer matters, showing the highest concentration in defamation cases, landlord and tenant disputes and employment matters. These are issues that require expedited review, given the time-sensitive nature of the cases.

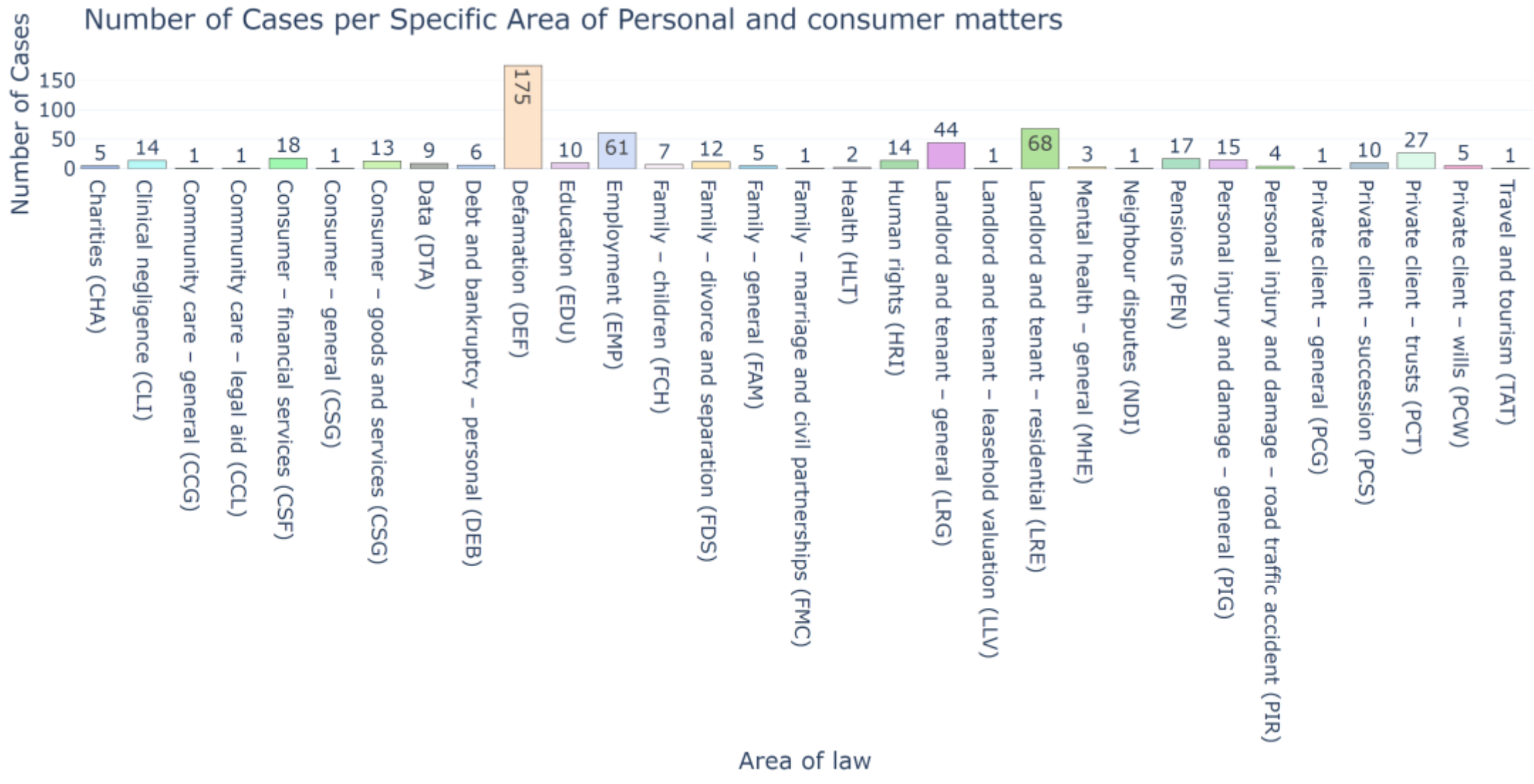

*Figure 7. Number of summary judgment cases per specific area of personal and consumer matters*

Finally, Figure 8 shows the distribution of cases within public law, with tax, planning and environmental issues being the most prominent.

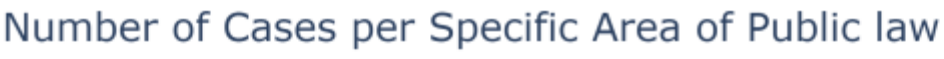

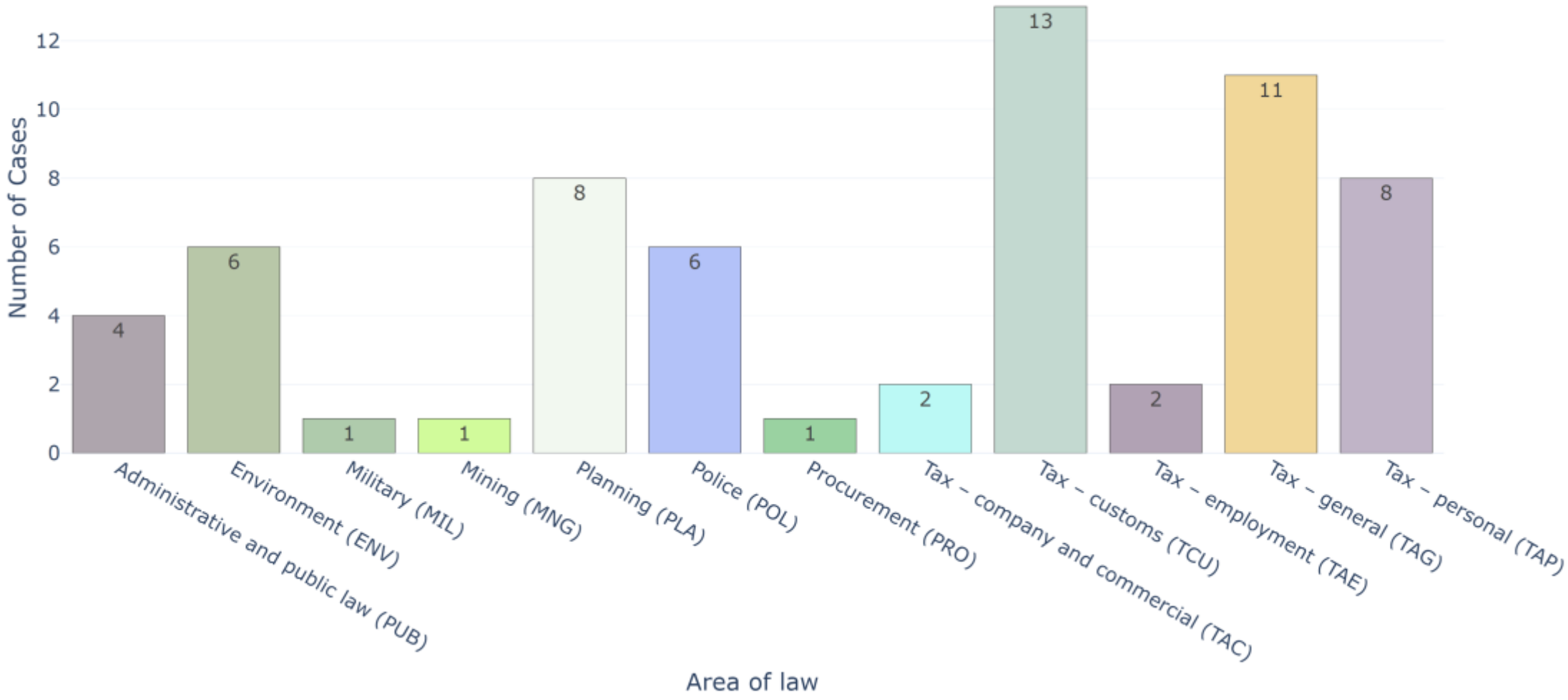

*Figure 8. Number of summary judgment cases per specific area of public law*

### *5.4.2 Distribution of topics over time*

In this section, we explore the temporal dynamics of summary judgment cases across various legal topics. The analysis is underpinned by visual representations that trace the frequency and distribution of cases across different periods. This temporal analysis is crucial for understanding trends, identifying emerging patterns and assessing the impact of legal and societal changes on the judicial processes. As explained in Section 4.1, the subset of summary judgment cases includes decisions between 26 April 1999 and 7 June 2023.

Figure 9 presents a line graph depicting the annual distribution of summary judgment cases within higher-level areas of law. This shows fluctuations and trends, including the predominance of commercial and dispute resolution law. It also shows a large spike in summary judgment cases that coincides with the COVID-19 pandemic.

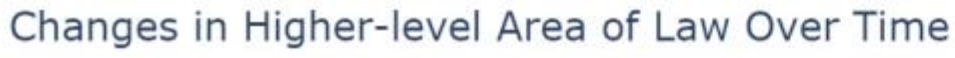

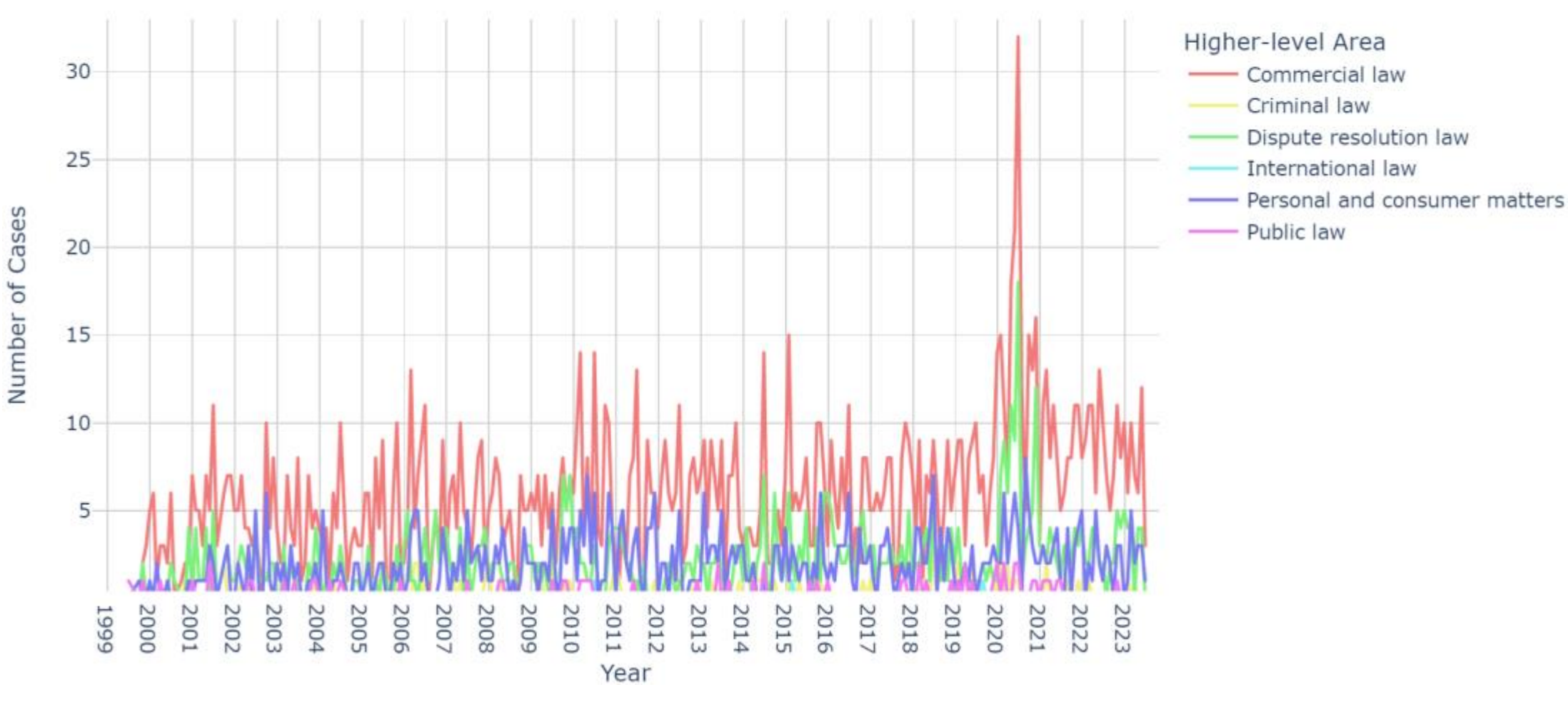

*Figure 9. Distribution of higher-level areas of law across time*

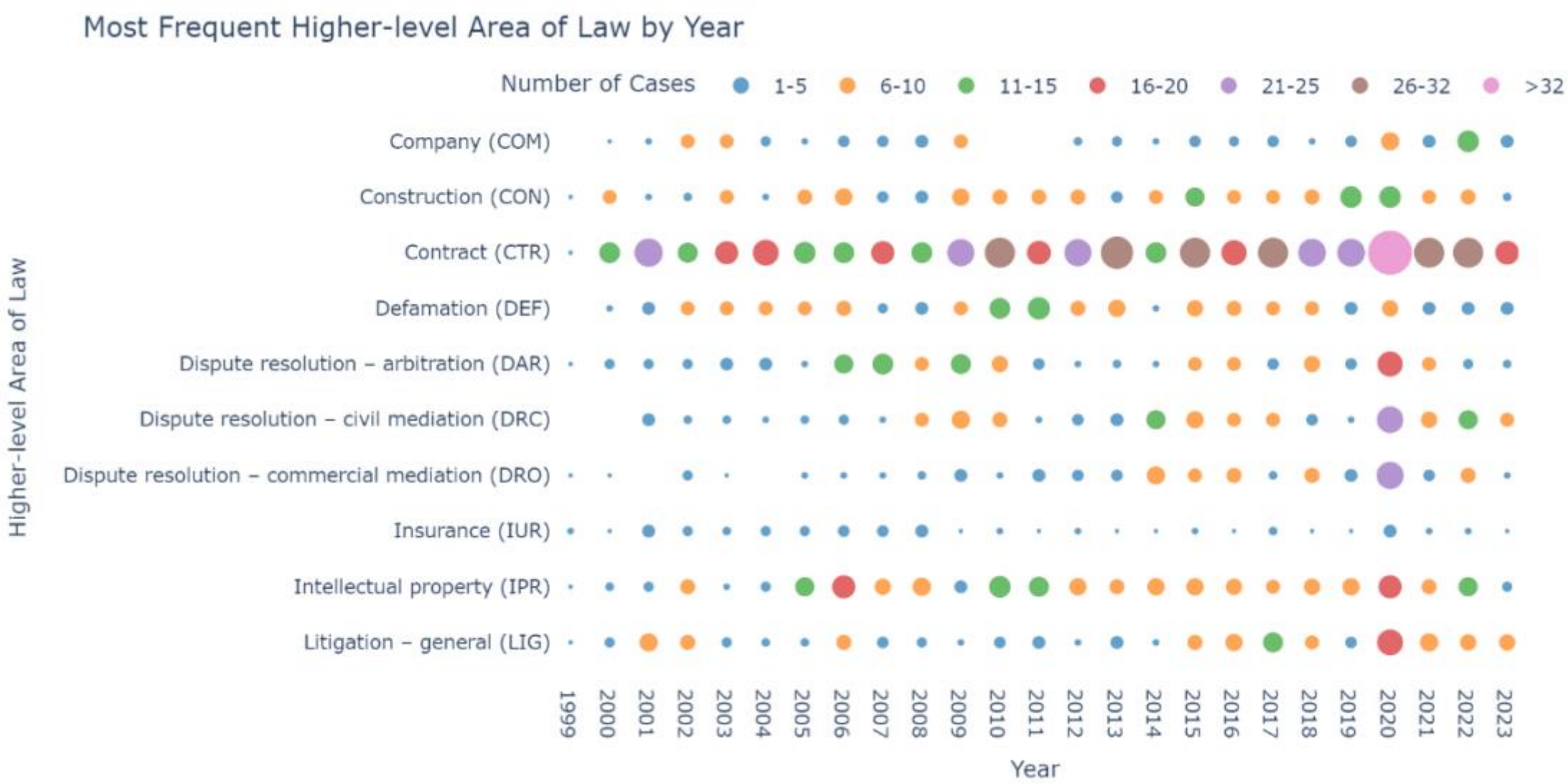

Figure 10 focuses on the top areas of law, plotting their distribution across time to highlight which specific topics have dominated the landscape of summary judgments at different times.

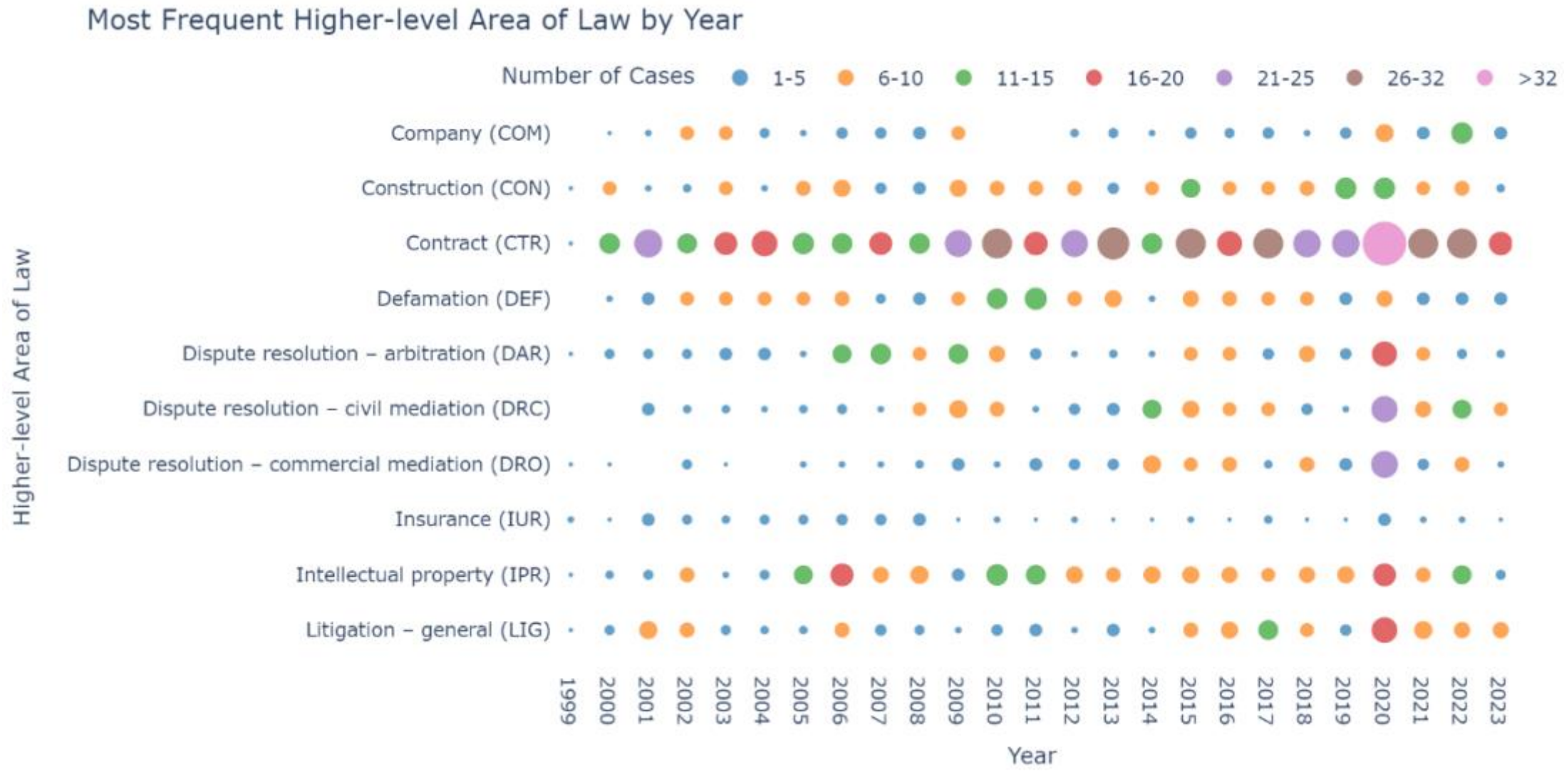

*Figure 10. Distribution of the most frequent cases of law across time. The size of the circle offers a quick guide to the relative number and colour coding shows the number range of topics per year.*

The temporal analysis of summary judgment cases across various legal topics provides a dynamic overview of how the legal landscape adapts over time. Further details are discussed in Section 6.1.

### *5.4.3 Distribution of topics across courts*

This subsection examines the distribution of summary judgment cases across the courts. Figure 11 shows the number of summary judgment cases across all UK courts in the dataset. This reveals the top five courts for summary judgment cases are the England and Wales High Court, including the Chancery Division, Queen's Bench Division (now King's Bench Division), Commercial Court, Technology and Construction Court, and Patents Court, and the England and Wales Court of Appeal (Civil Division).

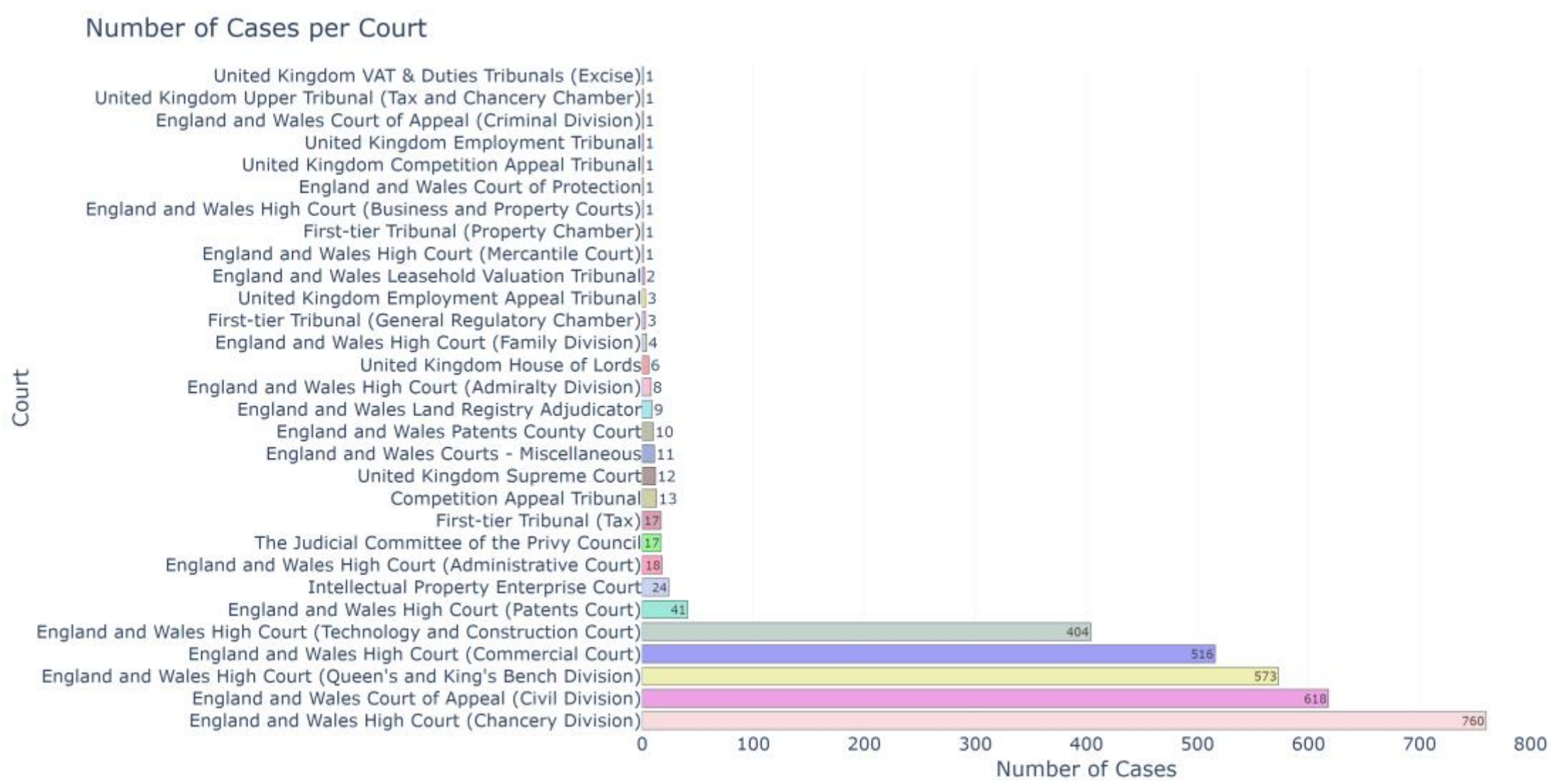

*Figure 11. Number of summary judgment cases per court*

Figure 12 to Figure 17 show the frequency of specific areas within each court. It is reflective of the specialisation of each court.

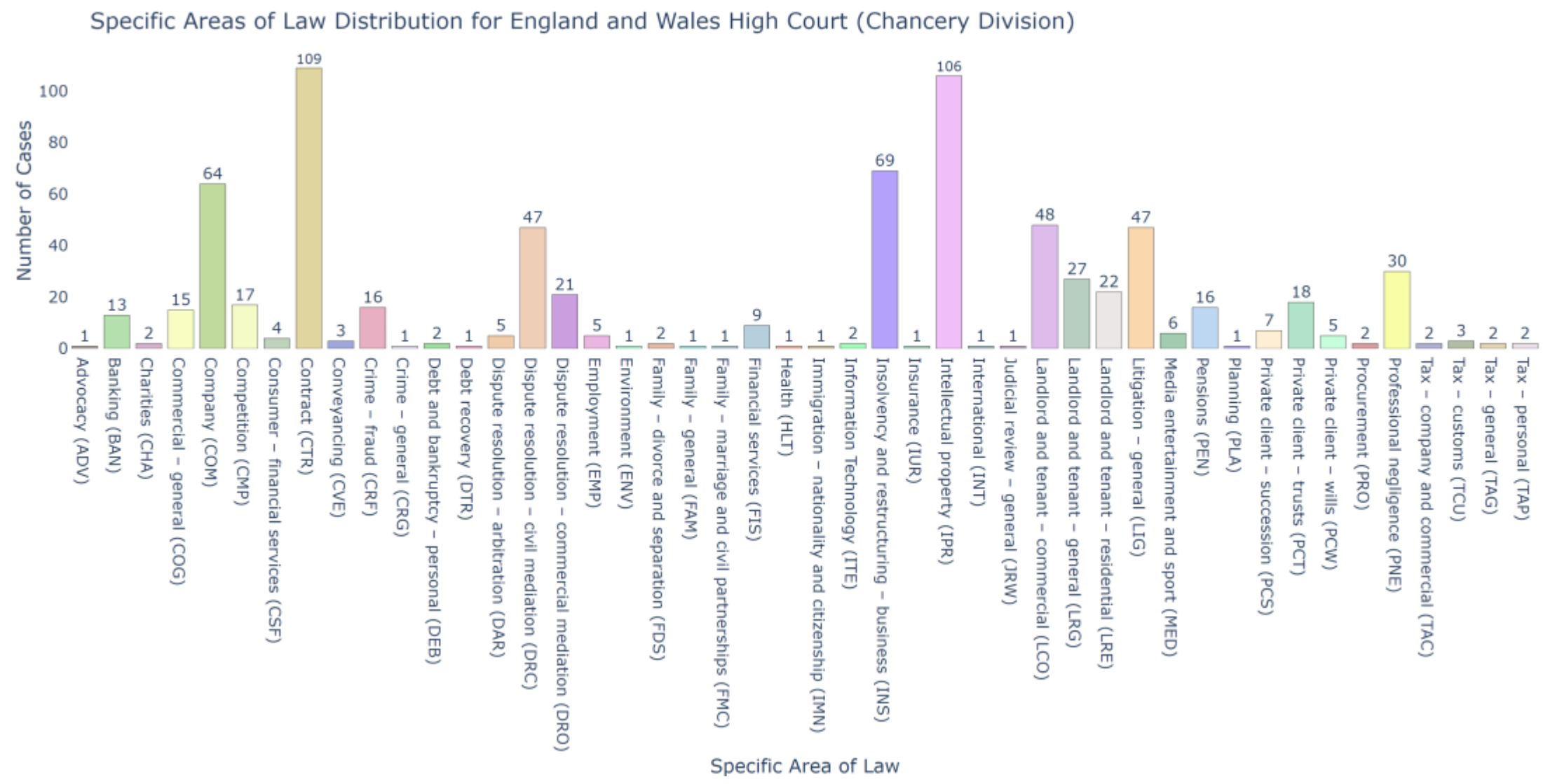

*Figure 12. Number of specific areas of law for England and Wales High Court (Chancery Division)*

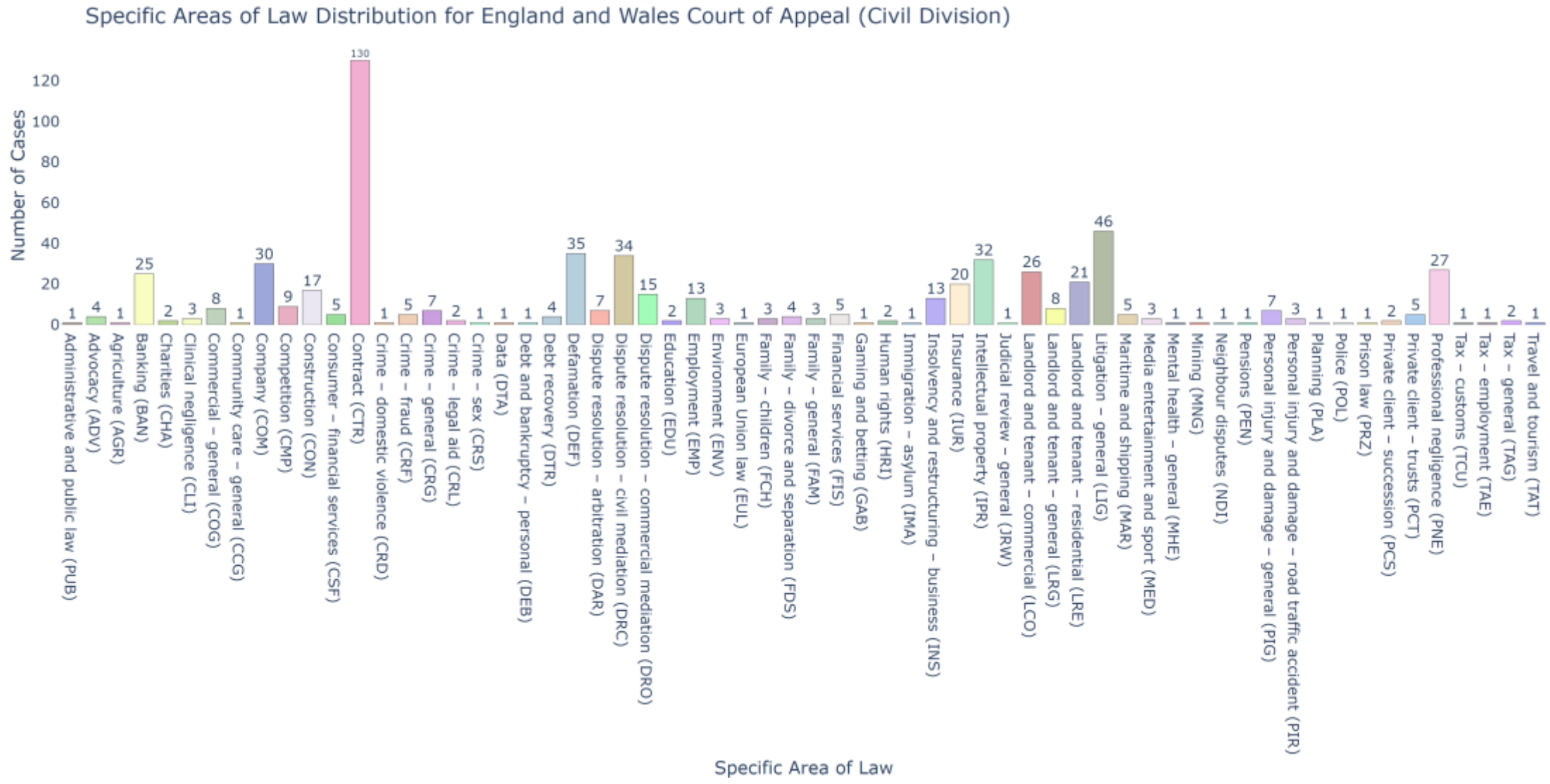

*Figure 13. Number of specific areas of law for England and Wales Court of Appeal (Civil Division)*

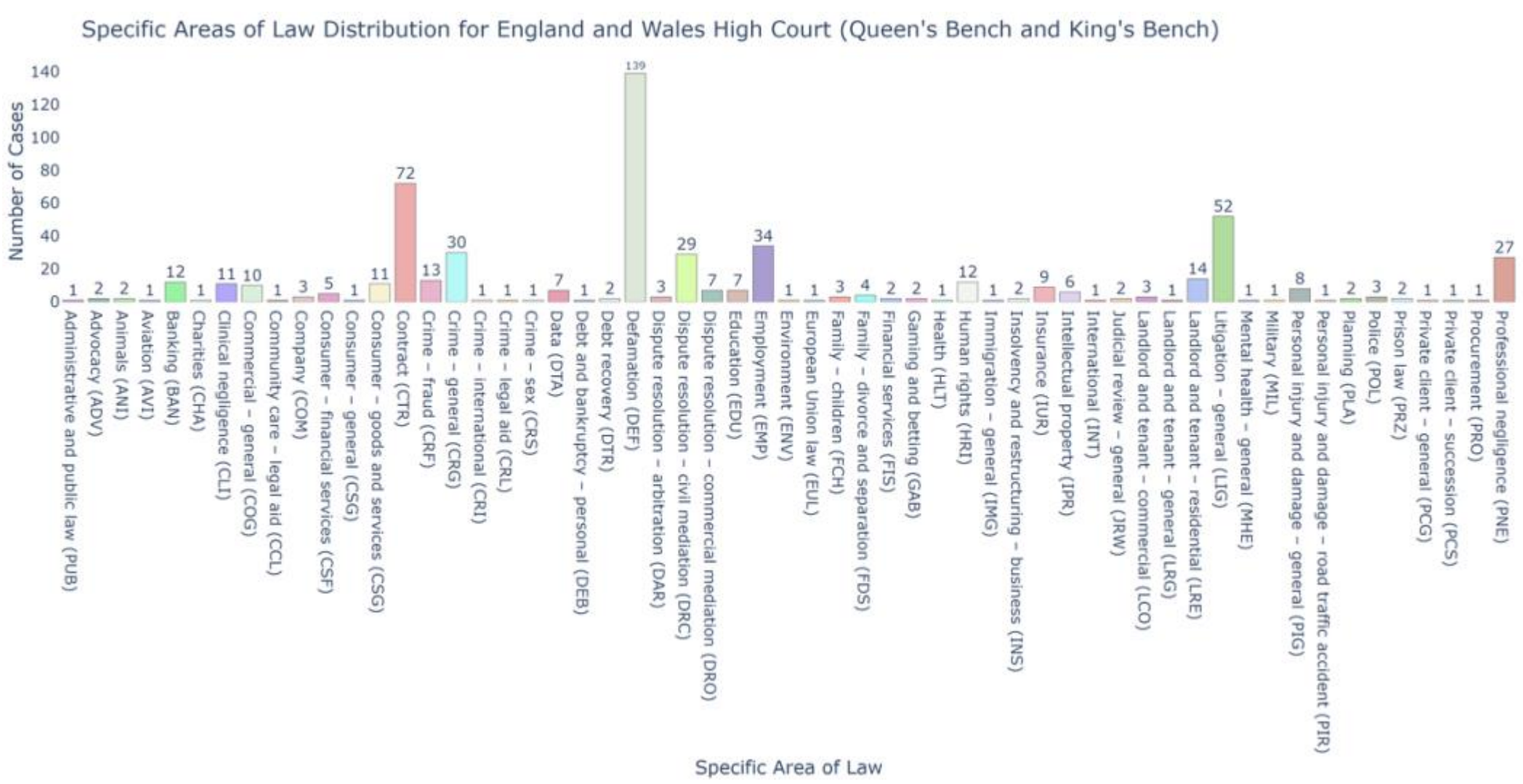

*Figure 14. Number of specific areas of law for England and Wales High Court (Queen's Bench and King's Bench)*

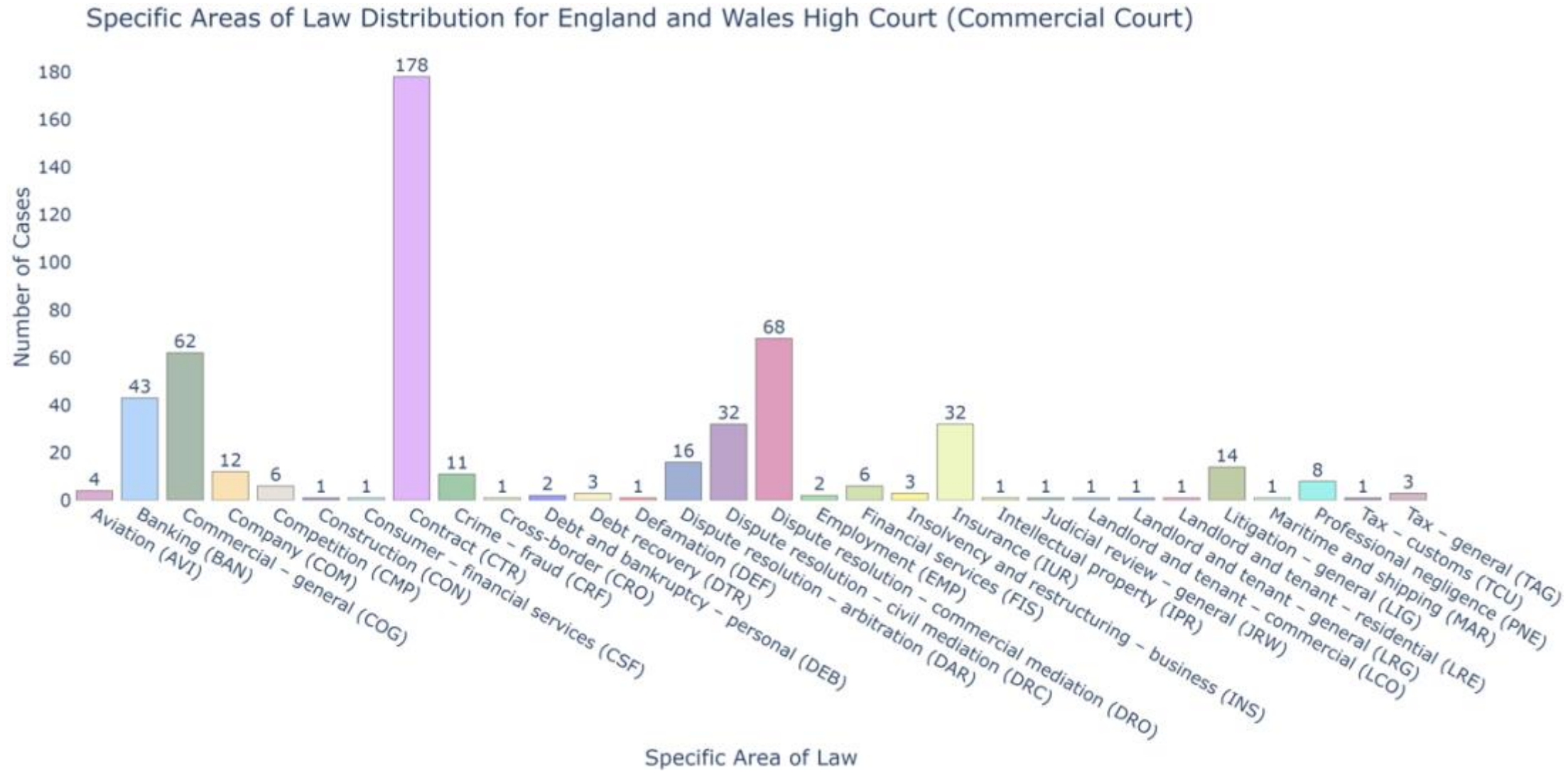

*Figure 15. Number of specific areas of law for England and Wales High Court (Commercial Court)*

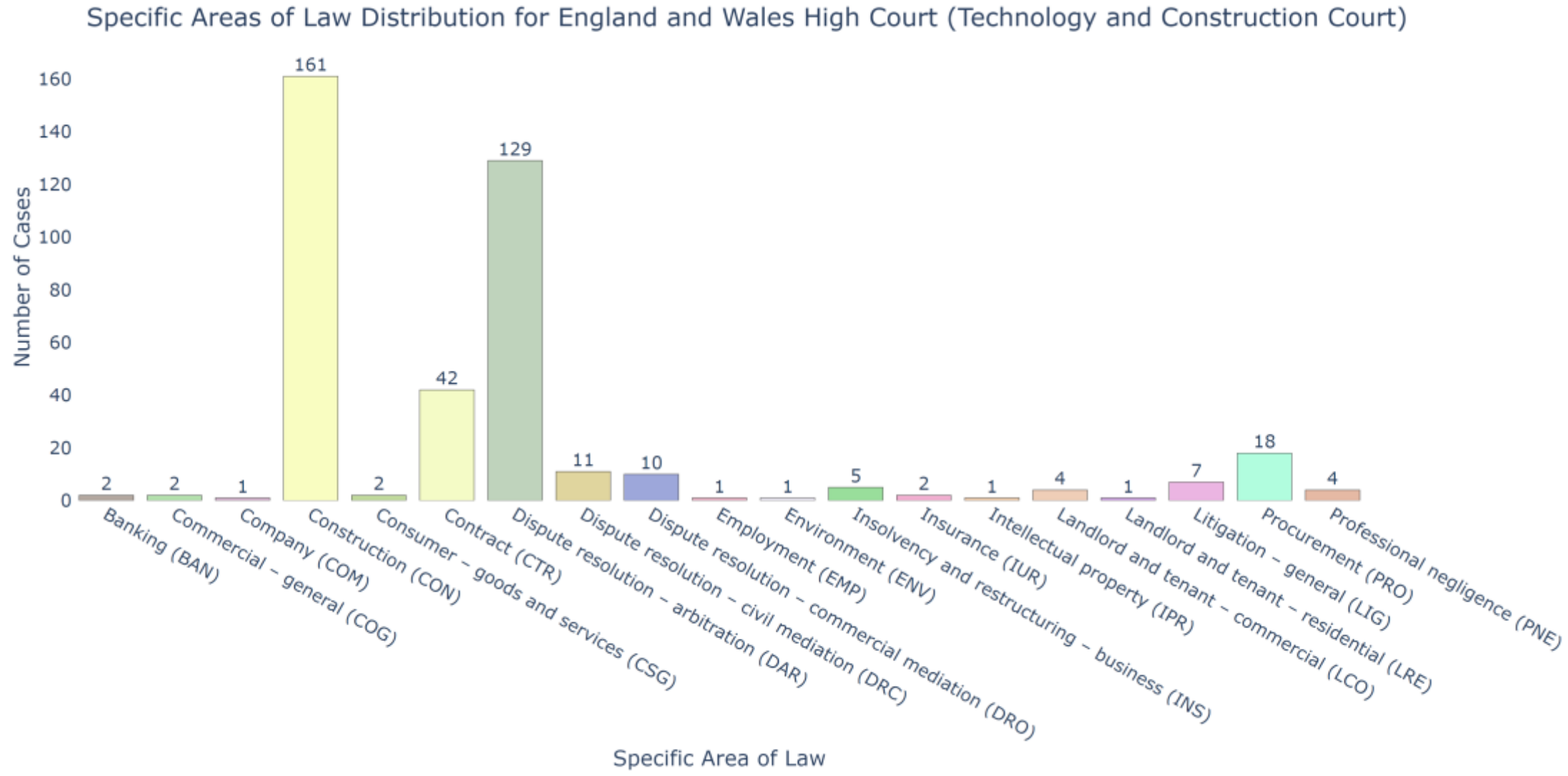

*Figure 16. Number of specific areas of law for England and Wales High Court (Technology and Construction Court)*

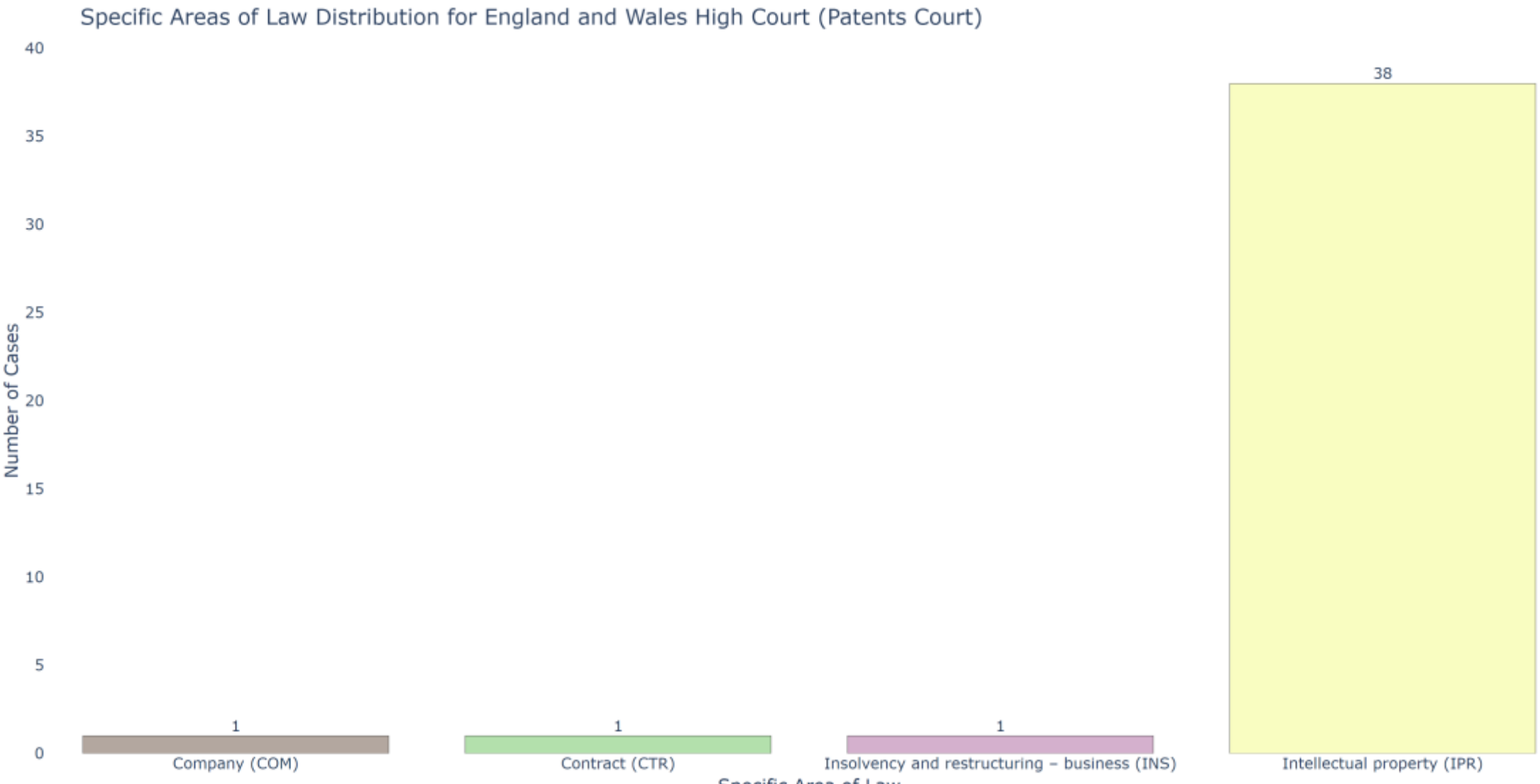

*Figure 17. Number of specific areas of law for England and Wales High Court (Patents Court)*

The visual analyses presented in Section 5.4 provide a comprehensive overview of the distribution and dynamics of summary judgment cases within the UK legal system, offering detailed insights into the prevalence and characteristics of these cases across different legal domains and courts. We now move into a more detailed discussion of these trends and insights into the topics of summary judgment cases.

## 6 **Discussion**

### 6.1 *Landscape of summary judgment case law*

Our findings illuminate the landscape of summary judgment cases in UK courts since its last major reform in 1999, revealing distinct patterns and trends across different legal domains and historical periods. The utilisation of summary judgments has fluctuated in response to various factors, including legislative reforms, changes in procedural rules and broader socio-economic developments.

Our results indicate a significant predominance of commercial law in summary judgment applications, as shown in Figure 2. This trend underscores the dominant role of summary judgments in resolving commercial disputes efficiently. Figure 3, in particular, shows the highest concentration of commercial law cases in contracts, followed by intellectual property, construction, company, banking, and professional negligence. Commercial disputes often involve significant financial stakes and, therefore, benefit from faster judicial processing to minimise business disruption and economic losses. Commercial law, including many contractual law issues, requires certainty and procedural simplicity, which heavily influences the frequent application of summary judgments in these areas (Halligan 2022). The construction of contractual documents, including issues of implied terms or contract variations, should be decided quickly through summary judgment, provided it is relatively straightforward (Sime 2022, 281).

Court efficiency is crucial in maintaining the flow of commerce and avoiding the economic stagnation that prolonged litigation can cause (Halligan 2022). This is particularly pertinent in sectors like construction, where project timelines and financing arrangements can be sensitive to delays. The England and Wales High Court (Technology and Construction Court) has “enthusiastically supported the use of applications for summary judgment to enforce the decisions of adjudicators” (Redmond 2001, 181). In particular, the summary judgment procedure allows an expedited timetable to resolve construction disputes and also shows the overlap with the dispute resolution topics.

Figure 5 shows the use of summary judgments in resolving traditional litigation matters and alternative dispute resolution mechanisms. Arbitration procedures have been criticised as inefficient and expensive, and so the adoption of summary judgment procedures is one way to improve the practice (Chong and Primrose 2017).

Figure 7 shows that defamation is the most frequent summary judgment topic within personal and consumer matters. Defamation cases often require prompt resolution (May 1985). The potential immediate impact on reputation, risks of wider dissemination of defamatory statements and prevention of ongoing harm or loss often lead to the use of expedited procedures or remedies such as injunctive relief. Areas of personal injury are often complex with disputed factual issues, so it is rare for a court to award summary judgment for general negligence claims. It is also true that a court will not normally award summary judgment in favour of a lawyer who is sued for professional negligence (Sime 2022),[5] although this area of law arises more frequently than more complex personal injury cases (compare Figure 3 and Figure 7).

Although summary judgment is not permitted in criminal trials due to the need for a full evaluation of evidence and a jury's determination of guilt, there is a presence of criminal-related topics in summary judgment cases within civil contexts (see Figure 4). When a defendant is convicted of a relevant offence, the civil courts may rely on this conviction to influence the adjudication of related civil claims.[6] This essentially means the criminal conviction can be used as evidence in civil proceedings unless significant proof to the contrary is presented. However, courts remain cautious in granting summary judgment when the underlying criminal conduct involves elements of deceit, dishonesty or other unlawful actions, as the complexity of these issues often requires a thorough examination of evidence, which is better suited to a trial setting.[7] An allegation of fraud can be a defence against a summary judgment application, particularly in the context of financial instruments like performance bonds. In scenarios involving claims of fraud, duress or illegality, the evidence must be particularly compelling to overcome the strong presumptions against summary judgment in such cases.[8] Such detail explains the high prevalence of the topic of fraud in summary judgment cases.

The data indicates the use of the summary judgment procedure in courts that handle complex commercial disputes, including the England and Wales High Court divisions including the King's & Queen's Bench, Chancery and England and Wales Court of Appeal (Civil), as well as specialised courts including the Technology and Construction Court, Commercial Court and Patents Court (Figure 11). The procedural use of summary judgments in these courts can be attributed to the nature of commercial litigation, which often involves clear-cut legal or factual questions that can be decided without a full trial (Sime 2022; Halligan 2022). This aligns with the broader judicial aim to streamline litigation processes and reduce unnecessary legal expenses and court time, especially in high-stakes commercial matters where swift resolution is often paramount. However, Phillips (2005) has identified some of the risks of summary judgment in these disputes. While summary judgment is increasingly invoked in intellectual property disputes to avoid the length and expense of a full trial, unlike full trials normally heard by a specialist intellectual property judge, applications for summary judgment are "frequently heard by judges in the Chancery Division who may have little experience of the area" (Phillips 2005). This disparity can be seen between the 106 intellectual property cases before the England and Wales High Court (Chancery Division) (Figure 12) and the 41 intellectual property cases before the England and Wales High Court (Patents Court) (Figure 17). Law and policymakers need to consider the balance between effective and swift administration of justice and the need for specialised judicial practice.

---

[5] *Sciortino v Beaumont* [2021] Ch 365.

[6] Civil Evidence Act 1968, s 11(2); *Dummer v Brown* [1953] 1 QB 710.

[7] *Espirit Telecoms UK Ltd v Fashion Gossip Ltd* (2000) LTL 27/7/00.

[8] *United Trading Corporation SA v Allied Arab Bank Ltd* [1985] 2 Lloyd's Rep 554; *Solo Industries UK Ltd v Canara Bank* [2001] 1 WLR 1800 at [32]; *Banque Saudi Fransi v Lear Siegler Services Inc* [2007] 1 All ER (Comm) 67 at [16].

A notable spike in summary judgment cases appeared in 2020, which highlights its usage in response to the extraordinary circumstances of the COVID-19 pandemic. The COVID-19 pandemic in 2020 brought unprecedented challenges to judicial systems worldwide, including a significant increase in caseloads and a consequent backlog within the UK court system (Select Committee on the Constitution 2021). This surge likely reflects the courts' agility in utilising summary judgments to expedite dispute resolution during a period when traditional court operations were disrupted. The increased use of summary judgments during the pandemic, however, also raises the question of whether summary judgments were handed down in cases that would have deserved a full trial.

### 6.2 *Topic Hallucinations*

Hallucinations present a significant challenge in the various applications of LLMs. LLMs compress vast amounts of training data into dense representations captured in the model's weights without storing verbatim text. This architecture predisposes LLMs to generate responses that approximate the context of a query but may inaccurately reflect the specifics of the training data, particularly under compression techniques that simplify the model's weights (Xu et al. 2024).

Despite recent interest in using LLMs in topic modelling (Wang et al. 2023; Xu et al. 2023), literature on effectively mitigating hallucinated topics remains largely unexplored. As outlined in Section 4.3, we have adapted some strategies from recent literature on managing hallucinations with LLMs.

We implemented a feedback mechanism whereby the LLM evaluates its own responses. To this end, we introduced a feedback mechanism designed to encourage the LLM to critically evaluate its own reasoning processes prior to providing the final classification of a legal area. Often referred to as "recall prompting" or "self-reflection," this method requires the model to articulate its rationale and assign a confidence score before delivering its final prediction. Emerging research demonstrates that prompting LLMs to justify or reconsider their initial responses can lead to improved accuracy (Regan et al. 2024; Hanqi et al. 2024). The underlying theory suggests that, by engaging in a self-reflective step, the model gains an opportunity to detect and correct logical, mathematical, or inferential errors that may have arisen in the initial phase of response generation (Renze et al. 2024). In essence, this practice functions analogously to a human expert double-checking work for errors, potentially enhancing performance on complex tasks. Without such reflective mechanisms, LLMs may resort to heuristic strategies, such as majority voting—where the model's final answer simply reflects the most frequently suggested responses across multiple iterations. Such aggregation strategies alone, even though frequently employed in practice (Li et al. 2024b), may be insufficient when dealing with nuanced legal topics that demand high precision. By contrast, self-reflection and iterative refinement appear to surpass traditional scalar reward approaches, fostering a recursive improvement cycle where outputs are continuously generated, critiqued, and optimized (Lu et al. 2023).

Drawing from the work of Mu et al. (2024b), training a model to discern between relevant and irrelevant or hallucinated topics can be achieved based on direct feedback. By integrating this strategy, an LLM is systematically guided to align its outputs with predefined criteria, enhancing the accuracy of its topic classification. Although the application of such feedback mechanisms in legal area classification is novel, the principle of using feedback to refine LLM outputs has been implemented in Shinn et al. (2023). Yan and Xu (2023) developed an iterative optimisation framework that enhances LLM responses by incorporating evaluative prompts. Shinn et al. (2023) introduced a 'reflection' strategy, where an LLM reviews and adjusts its answers based on previous outputs, fostering a self-corrective process.

These techniques are akin to instruction-tuning, where language models are fine-tuned on data described through specific instructions to improve their task performance (Wei et al., 2022). Our application of 'closed-set prompting'—a method where the LLM is presented with a predefined set of potential answers and selects the most appropriate one—is another form of this approach.

The ability of LLMs to comprehend and analyse complex, multidimensional legal texts offers a significant advantage over traditional methods. As Mu et al. (2024b) highlight, the transformative capabilities of LLMs extend beyond simple categorisation to grasp contextual nuances and thematic subtleties, allowing for a more detailed and nuanced classification of legal topics. However, ensuring the reliability of these models in legal applications also necessitates incorporating robust mechanisms to counteract hallucinations.

Although a contextual or theoretically limited taxonomy may present challenges to legal topic classification (Grajzl and Murrell, 2022), our work to develop a functional and comprehensive legal taxonomy focused on UK courts has resulted in promising accuracy. Continued research into feedback mechanisms and reflective strategies will be crucial in advancing the practical applications of LLMs in legal area classification, reducing the occurrence of hallucinations, and enhancing the overall trustworthiness of topic-classifying legal corpora.

### 6.3 *Limitations*

While our study offers significant insights into the legal area classification of summary judgment cases using LLMs and a new legal taxonomy for UK law, it is not without limitations.

First, our analysis did not include a comparison with other language models such as GPT, Llama or Gemini, nor traditional topic modelling approaches such as LDA, NMF or LSA. While the LLM Claude 3 Opus showed promising results with an overall accuracy of 88.30%, the absence of a comparative analysis limits our ability to contextualise these findings within the broader landscape of topic modelling techniques. Future research could benefit from a comparative study to evaluate the relative strengths and weaknesses of various models in the context of legal document analysis.

Second, our study was based on a relatively small sample size of 3078 summary judgment cases. Although this dataset is substantial within the specific context of UK summary judgments, it may not fully capture the diversity and complexity of all legal cases. A larger dataset could provide more robust insights and allow for more generalisable conclusions. Additionally, expanding the dataset to include cases from other jurisdictions or different types of legal documents could enhance the applicability of our findings.

Third, our methodology involved the manual evaluation of a subset of cases to verify the accuracy of the LLM's topic classifications. While this approach was necessary to assess the model's performance, it introduces a degree of subjectivity. The evaluation process relied on the judgment of legal experts, which, although informed, is inherently subjective and may vary among different evaluators. Future studies could aim to develop more objective and automated evaluation metrics to complement human assessments.

Finally, there is an inherent challenge of information leakage in the context of LLMs, particularly commercial LLMs where we lack oversight of the training data (Sargeant and Magnusson 2024; Sarkar and Vafa 2024). In the case of legal information, many online summaries, case notes or even html versions of the case judgments could be part of the training dataset. However, even where such information is available it is unlikely that true contamination would occur because there would not be any specific links to our predefined taxonomy of legal areas. This makes it unlikely that the model is simply regurgitating existing classifications from external sources.

In any event, it is not something that can be easily tested for a variety of reasons. First, providing a case name or citation to the model is not neutral information. It may indeed draw from training data that includes details about the case, but it could equally inform the model simply based on the case name or citation itself. For example, a neutral citation for the Employment Appeals Tribunal (e.g., [2024] EAT 15) would exclusively relate to "Employment (EMP)", and neutral citation to the United Kingdom VAT & Duties Tribunals (Customs) (e.g., [2009] UKVAT(Customs) C00267) are all related to Tax – customs (TCU). This is true for a large number of the UK courts and tribunals that are divided by specialty. Additionally, certain case names contain explicit clues about the legal area, such as "*C6, R (on the application of) v Secretary of State for the Home Department (asylum seekers' permission to work)* [2021] UKUT 94 (IAC)" clearly links to Immigration – asylum (IMA), or "*Davis (As Trustee In Bankruptcy of Jackson) v Jackson & Anor* [2017] EWHC 698 (Ch)" naturally falling under Debt and bankruptcy – personal (DEB). The risks of information leakage are important to consider in the context of future legal applications of LLMs.

In summary, while our study contributes to the field of computational legal studies by demonstrating the potential of LLMs in legal area classification, it also highlights the need for further research to address these limitations. Comparative analyses with other models, larger and more diverse datasets, objective evaluation methods, and testing across different types of legal cases are recommended to build on the findings of this study.

## 7 Conclusion

This study has successfully applied emerging techniques for LLM-based legal area classification to UK summary judgment cases. By developing and applying a functional taxonomy, we have enhanced the precision of topic classification, offering a nuanced view of legal topics and their applications in summary judgments. Our research highlights the robust potential of LLMs in handling complex legal texts and providing accurate classifications, achieving an accuracy of 87.13% and, when adjusted for minor errors, an accuracy of 88.20%.

The findings of this study reveal significant trends across different legal domains, notably in commercial and dispute resolution laws, and underscore the prevalent use of summary judgments in these areas. These insights are crucial for legal professionals and policymakers aiming to optimise judicial processes and balance the efficiency of legal proceedings with the fairness and accessibility of justice. Our research also underscores the ongoing challenges and limitations associated with topic classification in the legal field, particularly the occurrence of topic hallucinations and the nuanced understanding required to interpret legal language correctly.

Looking forward, this work lays a solid foundation for further exploration and refinement of computational methods in legal analytics. The proposed methodology and the developed taxonomy of legal areas provide a framework that can be extended to other areas of law, potentially revolutionising how legal data is processed and understood. Our study not only sheds light on the dynamics of summary judgment usage in UK courts but also underscores the importance of continuous evaluation of procedural laws to balance the goals of efficiency and justice in the legal system.

## Appendix A: Prompt

*A.1 Final Prompt*

"Dear Claude Opus,

I would like you to carefully review the details of a legal case that has been filed by a judge. After analyzing the facts and legal issues involved, please determine the most appropriate primary topic that the case falls under, based only on the following comprehensive list of legal topics:

{", ".join(legal_topics)}

If the case involves multiple legal issues, please identify one primary topic that is most central and relevant to the case. Avoid selecting a topic that is only tangentially related or a peripheral issue. The topic description should provide sufficient information for a judge to properly classify the case.

After determining the primary topic, please consider whether the case can also be categorized under any secondary topics from the legal topics list above only. If so, include the most relevant secondary topic. If there are no applicable secondary topics, simply state 'none'.

Before providing your response, please ensure that the selected primary and secondary topics are actually present in the list of legal topics provided above. If the topics you have chosen are not found in the list, please reconsider your selection and choose topics that are included in the list that meet the description of the primary and secondary topic.

Please format your response as follows:

Primary topic: [topic 1]

Secondary topic: [topic] or none

In your response, please provide a brief explanation (around 10 words) of your reasoning and cite any relevant legal principles or precedents that support your decision. Use the format: [Explanation: ]. Additionally, include a confidence score (1-5, with 5 being the most confident) for your topic allocation. Use the format: [Confidence score: ]

Thank you for your thorough analysis and attention to detail in this matter.

Case details:

{case}"

*A.2 Developing Prompt with Claude 3 Opus*

**User:**

I want you to write a Claude Opus custom made prompt.

The prompt is to consider a case filled by a judge. Then it is to allocate it to a specific topic.

The list of topics are given below:

[the full list was inserted here, each topic was pre-fixed with a number from 1 to 6 indicating which group it belonged to in terms of wider area of law. A carriage return after each topic was used, for example:

1 Administrative and public law (PUB)

2 Advocacy (ADV)

3 Agriculture (AGR)

3 Animals (ANI)]

Please note that topics that have common numerals indicate that they share the same general topic of law. I want you to craft a prompt to ask Claude Opus whether a case belongs to any one of the topics. then ask it again if the same case can be categorized in another secondary topic. For example, the case may be primary about two topics and not just one. You must avoid selecting topics that are not central to the case, i.e., a good topic description would be sufficient information to properly classify a case by a judge.

Output the response by giving the primary topics (without a number) followed by a comma then the secondary topic if there is one. If there is no secondary topic, simply output the word 'none'.

please format the prompt to be legally correct in language too.

The prompt must list all of the above areas of law in it, and the topic selection must come from it and from nowhere else.

Once you have selected the topic, perform a form of test to ensure it is correct, suggest a method please.

**Claude 3 Opus**:

"Dear Claude Opus,

I would like you to carefully review the details of a legal case that has been filed by a judge. After analyzing the facts and legal issues involved, please determine the most appropriate primary topic(s) that the case falls under, based on the following comprehensive list of legal topics:

Administrative and public law, Advocacy, Agriculture, Animals, Aviation, Banking, Benefits and allowances, Business - general, Business - finance, Charities, Clinical negligence, Commercial – general, Commercial – property, Community care – general, Community care – legal aid, Company, Competition, Construction, Consumer – general, Consumer – financial services, Consumer – goods and services, Consumer – utilities, Contract, Conveyancing, Corporate finance, Crime – general, Crime – domestic violence, Crime – fraud, Crime – international, Crime – juvenile, Crime – legal aid, Crime – sex, Cross-border, Data, Debt and bankruptcy – personal, Debt recovery, Defamation, Dispute resolution – arbitration, Dispute resolution – civil mediation, Dispute resolution – commercial mediation, Education, Elderly care – general, Elderly care – care home fees, Employment, Energy and natural resources, Environment, European Union law, Family – general, Family – children, Family – divorce and separation, Family – legal aid, Family – marriage and civil partnerships, Family – mediation, Family – mediation – legal aid, Family – public law – legal aid, Financial services, Gaming and betting, Health, Human rights, Immigration – general, Immigration – asylum, Immigration – legal aid, Immigration – nationality and citizenship, Information Technology, Insolvency and restructuring – business,

Insurance, Intellectual property, International, Judicial review – general, Judicial review – legal aid, Landlord and tenant – general, Landlord and tenant – commercial, Landlord and tenant – leasehold valuation, Landlord and tenant – residential, Lasting power of attorney, Litigation – general, Litigation – small claims, Litigation – tickets and fines, Maritime and shipping, Media entertainment and sport, Mental health – general, Mental health – legal aid, Mergers and acquisitions, Military, Mining, Motoring, Neighbour disputes, Pensions, Personal injury and damage – general, Personal injury and damage– road traffic accident, Planning, Police, Prison law, Private client – general, Private client – international, Private client – succession, Private client – trusts, Private client – wills, Procurement, Professional negligence, Tax – general, Tax – company and commercial, Tax – customs, Tax – employment, Tax – money and debt, Tax – personal, Travel and tourism.

If the case involves multiple legal issues, please identify up to two primary topics that are most central to the case. Avoid selecting topics that are only tangentially related. The topic description should provide sufficient information for a judge to properly classify the case.

After determining the primary topic(s), please consider whether the case can also be categorized under any secondary topics from the list above. If so, include the most relevant secondary topic. If there are no applicable secondary topics, simply state 'none'.

Please format your response as follows:

Primary topic(s): [topic 1], [topic 2 if applicable]

Secondary topic: [topic] or none

To ensure the accuracy of your topic allocation, please briefly explain your reasoning and cite any relevant legal principles or precedents that support your decision.

Thank you for your thorough analysis and attention to detail in this matter.

Sincerely,

[Your name]"

**User:**

Is there anything I am missing to improve this?

**Claude 3 Opus**:

Your prompt is quite comprehensive and well-structured. However, there are a few minor adjustments you could make to further improve it:

1. Clarify the output format: Specify that Claude Opus should provide the primary and secondary topics without their corresponding numbers to avoid any confusion.

2. Emphasize the importance of selecting the most relevant topics: Stress that the selected topics should be the most central and relevant to the case, and that peripheral issues should not influence the topic allocation.

3. Request a confidence score: Ask Claude Opus to provide a confidence score (e.g., on a scale of 1-5) for its topic allocation, which could help assess the reliability of its decision.

4. Provide an example: Include a sample case and its corresponding topic allocation to serve as a reference point for Claude Opus.

5. Specify the response length: Indicate the desired length of Claude Opus's explanation for its topic selection to ensure a concise yet informative response.

Here's the revised prompt incorporating these suggestions:

"Dear Claude Opus,

I would like you to carefully review the details of a legal case that has been filed by a judge. After analyzing the facts and legal issues involved, please determine the most appropriate primary topic(s) that the case falls under, based on the following comprehensive list of legal topics:

[List of legal topics]

If the case involves multiple legal issues, please identify up to two primary topics that are most central and relevant to the case. Avoid selecting topics that are only tangentially related or peripheral issues. The topic description should provide sufficient information for a judge to properly classify the case.

After determining the primary topic(s), please consider whether the case can also be categorized under any secondary topics from the list above. If so, include the most relevant secondary topic. If there are no applicable secondary topics, simply state 'none'.

Please format your response as follows:

Primary topic(s): [topic 1], [topic 2 if applicable]

Secondary topic: [topic] or none

In your response, please provide a brief explanation (around 100 words) of your reasoning and cite any relevant legal principles or precedents that support your decision. Additionally, include a confidence score (1-5, with 5 being the most confident) for your topic allocation.

To ensure the accuracy of your topic selection, please refer to the following example:

Case: [Brief case description]
Primary topic(s): [topic 1], [topic 2 if applicable]
Secondary topic: [topic] or none
Explanation: [100-word explanation]
Confidence score: [1-5]

Thank you for your thorough analysis and attention to detail in this matter.

Sincerely,
[Your name]"

By incorporating these adjustments, the prompt should provide Claude Opus with clearer guidelines and expectations for its response, leading to more accurate and reliable topic allocations

## Appendix B: Taxonomy of Legal Areas

*B.1 Specific areas of law*

*Note: the number before each area of law indicates the higher-order area of law given below*

6 Administrative and public law (PUB)
3 Advocacy (ADV)
1 Agriculture (AGR)
1 Animals (ANI)
1 Aviation (AVI)
1 Banking (BAN)
5 Benefits and allowances (BEN)
1 Business - general (BUS)
1 Business - finance (BFI)
5 Charities (CHA)
5 Clinical negligence (CLI)
1 Commercial – general (COG)
1 Commercial – property (COP)
5 Community care – general (CCG)
5 Community care – legal aid (CCL)
1 Company (COM)
1 Competition (CMP)
1 Construction (CON)
5 Consumer – general (CSG)
5 Consumer – financial services (CSF)
5 Consumer – goods and services (CSG)
5 Consumer – utilities (CUT)
1 Contract (CTR)
1 Conveyancing (CVE)
1 Corporate finance (CFI)
2 Crime – general (CRG)
2 Crime – domestic violence (CRD)
2 Crime – fraud (CRF)
2 Crime – international (CRI)
2 Crime – juvenile (CRJ)
2 Crime – legal aid (CRL)
2 Crime – sex (CRS)
4 Cross-border (CRO)
5 Data (DTA)
5 Debt and bankruptcy – personal (DEB)
1 Debt recovery (DTR)
5 Defamation (DEF)
3 Dispute resolution – arbitration (DAR)
3 Dispute resolution – civil mediation (DRC)
3 Dispute resolution – commercial mediation (DRO)
5 Education (EDU)
5 Elderly care – general (ECG)
5 Elderly care – care home fees (ELH)
5 Employment (EMP)
1 Energy and natural resources (ENE)
6 Environment (ENV)
4 European Union law (EUL)
5 Family – general (FAM)

5 Family – children (FCH)
5 Family – divorce and separation (FDS)
5 Family – legal aid (FAL)
5 Family – marriage and civil partnerships (FMC)
5 Family – mediation (FME)
5 Family – mediation – legal aid (FML)
5 Family – public law – legal aid (FPL)
1 Financial services (FIS)
1 Gaming and betting (GAB)
5 Health (HLT)
5 Human rights (HRI)
4 Immigration – general (IMG)
4 Immigration – asylum (IMA)
4 Immigration – legal aid (IML)
4 Immigration – nationality and citizenship (IMN)
1 Information Technology (ITE)
1 Insolvency and restructuring – business (INS)
1 Insurance (IUR)
1 Intellectual property (IPR)
4 International (INT)
3 Judicial review – general (JRW)
3 Judicial review – legal aid (JRL)
5 Landlord and tenant – general (LRG)
1 Landlord and tenant – commercial (LCO)
5 Landlord and tenant – leasehold valuation (LLV)
5 Landlord and tenant – residential (LRE)
5 Lasting power of attorney (POA)
3 Litigation – general (LIG)
3 Litigation – small claims (LIS)
3 Litigation – tickets and fines (LIT)
1 Maritime and shipping (MAR)
1 Media entertainment and sport (MED)
5 Mental health – general (MHE)
5 Mental health – legal aid (MHL)
1 Mergers and acquisitions (MAA)
6 Military (MIL)
6 Mining (MNG)
6 Motoring (MOT)
5 Neighbour disputes (NDI)
5 Pensions (PEN)
5 Personal injury and damage – general (PIG)
5 Personal injury and damage– road traffic accident (PIR)
6 Planning (PLA)
6 Police (POL)
2 Prison law (PRZ)
5 Private client – general (PCG)
5 Private client – international (PCI)
5 Private client – succession (PCS)
5 Private client – trusts (PCT)
5 Private client – wills (PCW)
1 Procurement (PRO)
1 Professional negligence (PNE)
6 Tax – general (TAG)
6 Tax – company and commercial (TAC)
6 Tax – customs (TCU)

6 Tax – employment (TAE)
6 Tax – money and debt (TAM)
6 Tax – personal (TAP)
5 Travel and tourism (TAT)

*B.2 Higher-level areas of law*

1 Commercial law

Agriculture (AGR), Animals (ANI), Aviation (AVI), Banking (BAN), Business - general (BUS), Business - finance (BFI), Commercial – general (COG), Commercial – property (COP), Company (COM), Competition (CMP), Construction (CON), Contract (CTR), Conveyancing (CVE), Corporate finance (CFI), Debt recovery (DTR), Energy and natural resources (ENE), Financial services (FIS), Gaming and betting (GAB), Information Technology (ITE), Insolvency and restructuring – business (INS), Insurance (IUR), Intellectual property (IPR), Landlord and tenant – commercial (LCO), Maritime and shipping (MAR), Media entertainment and sport (MED), Mergers and acquisitions (MAA), Procurement (PRO), Professional negligence (PNE)

2 Criminal law

Crime – general (CRG), Crime – domestic violence (CRD), Crime – fraud (CRF), Crime – international (CRI), Crime – juvenile (CRJ), Crime – legal aid (CRL), Crime – sex (CRS), Prison law (PRZ)

3 Dispute resolution law

Advocacy (ADV), Dispute resolution – arbitration (DAR), Dispute resolution – civil mediation (DRC), Dispute resolution – commercial mediation (DRO), Judicial review – general (JRW), Judicial review – legal aid (JRL), Litigation – general (LIG), Litigation – small claims (LIS), Litigation – tickets and fines (LIT)

4 International law

Cross-border (CRO), European Union law (EUL), Immigration – general (IMG), Immigration – asylum (IMA), Immigration – legal aid (IML), Immigration – nationality and citizenship (IMN), International (INT)

5 Personal and consumer matters

Benefits and allowances (BEN), Charities (CHA), Clinical negligence (CLI), Community care – general (CCG), Community care – legal aid (CCL), Consumer – general (CSG), Consumer – financial services (CSF), Consumer – goods and services (CSG), Consumer – utilities (CUT), Data (DTA), Debt and bankruptcy – personal (DEB), Defamation (DEF), Education (EDU), Elderly care – general (ECG), Elderly care – care home fees (ELH), Employment (EMP), Family – general (FAM), Family – children (FCH), Family – divorce and separation (FDS), Family – legal aid (FAL), Family – marriage and civil partnerships (FMC), Family – mediation (FME), Family – mediation – legal aid (FML), Family – public law – legal aid (FPL), Health (HLT), Human rights (HRI), Landlord and tenant – general (LRG), Landlord and tenant – leasehold valuation (LLV), Landlord and tenant – residential (LRE), Lasting power of attorney (POA), Mental health – general (MHE), Mental health – legal aid (MHL), Neighbour disputes (NDI), Pensions (PEN), Personal injury and damage – general (PIG), Personal injury and damage– road traffic accident (PIR), Private client – general (PCG), Private client – international (PCI), Private client – succession (PCS), Private client – trusts (PCT), Private client – wills (PCW), Travel and tourism (TAT)

6 Public law

Administrative and public law (PUB), Environment (ENV), Military (MIL), Mining (MNG), Motoring (MOT), Planning (PLA), Police (POL), Tax – general (TAG), Tax – company and commercial (TAC), Tax – customs (TCU), Tax – employment (TAE), Tax – money and debt (TAM), Tax – personal (TAP)

*B.3 Amendments*

The following documents the amendments made based on the 'Areas of practice' list given by The Law Society.

Actions against the police – legal aid (APL) amended to Police (POL)
Added Animals (ANI)
Children (CHI) amended to Family – children (FCH)
Commercial litigation (COL) amended to Commercial – general (COG)
Commercial property (PCO) amended to Commercial – property (COP)
Added Community care – general (CCG)
Company / commercial – small businesses (COS) amended to Business (BUS)
Company and commercial (COM) amended to Company (COM)
Added Competition (CMP)
Consumer (CSU) amended to Consumer – general (CSG)
Conveyancing – residential (PRE) amended to Conveyancing (CVE)
Corporate finance (CFI) amended to Business - finance (BFI)
Added Crime – international (CRI)
Crime – motoring (CRM) amended to Motoring (MOT)
Added Data (DTA)
Added Dispute resolution – arbitration (DAR)
Elderly care (ELC) amended to Elderly care – general (ECG)
European Community law (EUN) amended to European Union law (EUL)
Family public law – legal aid (FPL) amended to Family – public law – legal aid (FPL)
Added Health (HLT)
Immigration – general (IMM) amended to Immigration – general (IMG)
Deleted Immigration – asylum – legal aid (IML)
Immigration – general – legal aid (IMG) amended to Immigration – legal aid (IML)
Added International (INT)
Judicial review (JRW) amended to Judicial review – general (JRW)
Added Landlord and tenant – general (LRG)
Added Landlord and tenant – leasehold valuation (LLV)
Licensing gaming and betting (LIC) amended to Gaming and betting (GAB)
Litigation – general (LIV) amended to Litigation – general (LIG)
Deleted LPA – Health and Welfare (LPH)
Deleted LPA – Property and Affairs (LPP)
Mental health (MHE) amended to Mental health – general (MHE)
Added Mining (MNG)
Personal injury (PIN) amended to Personal injury and damage – general (PIG)
Personal injury – RTA (PIR) amended to Personal injury and damage – road traffic accident (PIR)
Prison law – legal aid (PRZ) amended to Prison law (PRZ)
Added Private client - general (PCG)
Deleted Private client - disputed probate (PRP)
Deleted Private client - disputed trusts (PRT)
Deleted Private client - disputed wills (PRW)
Private client – probate (PCP) amended to Private client – succession (PCS)
Added Procurement (PRO)Tax (TAX) amended to Tax – general (TAG)
Added Tax – customs (TCU)
Deleted Tax – Employment/HR (TAH)

**Appendix C: Court Tiers**

| Court Category | Court Name |
|---|---|
| Tier 1:<br>Court of Last Resort | United Kingdom Supreme Court |
| | United Kingdom House of Lords |
| | The Judicial Committee of the Privy Council |
| Tier 2:<br>Appellate Court | England and Wales Court of Appeal (Civil Division) |
| | England and Wales Court of Appeal (Criminal Division) |
| Tier 2:<br>Appellate Tribunal | United Kingdom Employment Appeal Tribunal |
| | United Kingdom Competition Appeal Tribunal |
| Tier 3:<br>First Instance Court | England and Wales Court of Protection |
| | England and Wales Courts - Miscellaneous |
| | England and Wales High Court (Administrative Court) |
| | England and Wales High Court (Admiralty Division) |
| | England and Wales High Court (Business and Property Courts) |
| | England and Wales High Court (Chancery Division) |
| | England and Wales High Court (Commercial Court) |
| | England and Wales High Court (Family Division) |
| | England and Wales High Court (King's Bench Division) |
| | England and Wales High Court (Mercantile Court) |
| | England and Wales High Court (Patents Court) |
| | England and Wales High Court (Queen's Bench Division) |
| | England and Wales High Court (Technology and Construction Court) |
| | England and Wales Patents County Court |
| Tier 3:<br>First Instance Tribunal | England and Wales Land Registry Adjudicator |
| | England and Wales Leasehold Valuation Tribunal |
| | First-tier Tribunal (General Regulatory Chamber) |
| | First-tier Tribunal (Property Chamber) |
| | First-tier Tribunal (Tax) |
| | United Kingdom Employment Tribunal |
| | United Kingdom Upper Tribunal (Tax and Chancery Chamber) |
| | United Kingdom VAT & Duties Tribunals (Excise) |

**Appendix D: Accuracy per Area of Law**

| Area of Law | Precision | Recall | F1-score | Support | Accuracy | Total Cases |
|---|---|---|---|---|---|---|
| Advocacy (ADV) | 1.00 | 1.00 | 1.00 | 1 | 100.00% | 1 |
| Arbitration (ARB) | 1.00 | 1.00 | 1.00 | 2 | 100.00% | 2 |
| Aviation (AVI) | 1.00 | 1.00 | 1.00 | 1 | 100.00% | 1 |
| Banking (BAN) | 1.00 | 0.67 | 0.80 | 9 | 66.67% | 9 |
| Charities (CHA) | 1.00 | 1.00 | 1.00 | 1 | 100.00% | 1 |
| Clinical negligence (CLI) | 1.00 | 0.67 | 0.80 | 3 | 66.67% | 3 |
| Commercial - general (COG) | 1.00 | 1.00 | 1.00 | 10 | 100.00% | 10 |
| Company (COM) | 0.91 | 0.77 | 0.83 | 13 | 76.92% | 13 |
| Competition (CMP) | 1.00 | 1.00 | 1.00 | 6 | 100.00% | 6 |
| Construction (CON) | 1.00 | 0.92 | 0.96 | 24 | 91.67% | 24 |
| Consumer - goods and services (CSG) | 1.00 | 1.00 | 1.00 | 1 | 100.00% | 1 |
| Contract (CTR) | 1.00 | 0.77 | 0.87 | 65 | 76.92% | 65 |
| Conveyancing (CVE) | 1.00 | 1.00 | 1.00 | 3 | 100.00% | 3 |
| Crime - domestic violence (CRD) | 1.00 | 1.00 | 1.00 | 1 | 100.00% | 1 |
| Crime - fraud (CRF) | 1.00 | 0.33 | 0.50 | 3 | 66.67% | 3 |
| Crime - general (CRG) | 0.83 | 1.00 | 0.91 | 5 | 100.00% | 5 |
| Debt recovery (DTR) | 1.00 | 1.00 | 1.00 | 2 | 100.00% | 2 |
| Defamation (DEF) | 1.00 | 1.00 | 1.00 | 25 | 100.00% | 25 |
| Dispute resolution - arbitration (DAR) | 0.82 | 0.93 | 0.88 | 15 | 100.00% | 15 |
| Dispute resolution - civil mediation (DRC) | 0.16 | 1.00 | 0.27 | 3 | 100.00% | 3 |
| Dispute resolution - commercial mediation (DRO) | 0.08 | 1.00 | 0.14 | 1 | 100.00% | 1 |
| Education (EDU) | 1.00 | 1.00 | 1.00 | 2 | 100.00% | 2 |
| Employment (EMP) | 1.00 | 1.00 | 1.00 | 7 | 100.00% | 7 |
| Environment (ENV) | 1.00 | 1.00 | 1.00 | 1 | 100.00% | 1 |
| Family - general (FAM) | 1.00 | 1.00 | 1.00 | 1 | 100.00% | 1 |
| Financial services (FIS) | 1.00 | 0.80 | 0.89 | 5 | 80.00% | 5 |
| Human rights (HRI) | 0.00 | 0.00 | 0.00 | 1 | 0.00% | 1 |
| Information Technology (ITE) | 1.00 | 1.00 | 1.00 | 1 | 100.00% | 1 |
| Insolvency and restructuring – business (INS) | 1.00 | 0.85 | 0.92 | 13 | 84.62% | 13 |
| Insurance (IUR) | 1.00 | 0.78 | 0.88 | 9 | 77.78% | 9 |
| Intellectual property (IPR) | 1.00 | 1.00 | 1.00 | 20 | 100.00% | 20 |
| Landlord and tenant - commercial (LCO) | 0.91 | 1.00 | 0.95 | 10 | 100.00% | 10 |
| Landlord and tenant - general (LRG) | 0.40 | 1.00 | 0.57 | 2 | 100.00% | 2 |
| Landlord and tenant - residential (LRE) | 0.73 | 0.89 | 0.80 | 9 | 88.89% | 9 |
| Litigation - general (LIG) | 1.00 | 0.82 | 0.90 | 22 | 81.82% | 22 |
| Media entertainment and sport (MED) | 1.00 | 1.00 | 1.00 | 1 | 100.00% | 1 |

| Mental health - general (MHE) | 1.00 | 1.00 | 1.00 | 2 | 100.00% | 2 |
|---|---|---|---|---|---|---|
| Pensions (PEN) | 1.00 | 1.00 | 1.00 | 1 | 100.00% | 1 |
| Personal injury and damage – road traffic accident (PIR) | 1.00 | 1.00 | 1.00 | 1 | 100.00% | 1 |
| Planning (PLA) | 0.00 | 0.00 | 0.00 | 6 | 0.00% | 6 |
| Police (POL) | 1.00 | 1.00 | 1.00 | 1 | 100.00% | 1 |
| Privacy (PRI) | 1.00 | 1.00 | 1.00 | 1 | 100.00% | 1 |
| Private client - succession (PCS) | 1.00 | 1.00 | 1.00 | 2 | 100.00% | 2 |
| Private client - trusts (PCT) | 1.00 | 1.00 | 1.00 | 3 | 100.00% | 3 |
| Private client - wills (PCW) | 1.00 | 1.00 | 1.00 | 1 | 100.00% | 1 |
| Professional negligence (PNE) | 1.00 | 0.95 | 0.97 | 19 | 94.74% | 19 |
| Public procurement (PRO) | 1.00 | 0.50 | 0.67 | 2 | 50.00% | 2 |
| Tax - customs (TCU) | 1.00 | 1.00 | 1.00 | 2 | 100.00% | 2 |
| Tax - general (TAG) | 1.00 | 1.00 | 1.00 | 1 | 100.00% | 1 |
| Trusts (TRU) | 1.00 | 1.00 | 1.00 | 2 | 100.00% | 2 |